\PassOptionsToPackage{table}{xcolor}
\documentclass[pdflatex,sn-mathphys-num]{sn-jnl}

\usepackage{graphicx}%
\usepackage{multirow}%
\usepackage{amsmath,amssymb,amsfonts}%
\usepackage{amsthm}%
\usepackage{mathrsfs}%
\usepackage[title]{appendix}%
\usepackage{xcolor}%
\usepackage{textcomp}%
\usepackage{manyfoot}%
\usepackage{booktabs}%
\usepackage{algorithm}%
\usepackage{algorithmicx}%
\usepackage{algpseudocode}%
\usepackage{listings}%
\usepackage{makecell}
\usepackage[table]{xcolor}
\definecolor{lightgray}{gray}{0.95}
\usepackage{chngcntr}  
\newcommand{\supportinginfo}{
  \setcounter{section}{0}
  \renewcommand{\thesection}{S\arabic{section}}

  \setcounter{figure}{0}
  \renewcommand{\thefigure}{S\arabic{figure}}

  \setcounter{table}{0}
  \renewcommand{\thetable}{S\arabic{table}}

  \setcounter{equation}{0}
  \renewcommand{\theequation}{S\arabic{equation}}
}

\theoremstyle{thmstyleone}%
\theoremstyle{thmstyletwo}%

\theoremstyle{thmstylethree}%

\begin{document}

\title[A Foundation Model for Material Fracture Prediction]{A Foundation Model for Material Fracture Prediction}

\author*[1]{\fnm{Agnese} \sur{Marcato}}\email{amarcato@lanl.gov}

\author[1]{\fnm{Aleksandra} \sur{Pachalieva}}
\author[2]{\fnm{Ryley} \sur{G. Hill}}
\author[1]{\fnm{Kai} \sur{Gao}}
\author[2]{\fnm{Xiaoyu} \sur{Wang}}
\author[2]{\fnm{Esteban} \sur{Rougier}}
\author[2]{\fnm{Zhou} \sur{Lei}}
\author[3]{\fnm{Vinamra} \sur{Agrawal}}
\author[3]{\fnm{Janel} \sur{Chua}}
\author[1]{\fnm{Qinjun} \sur{Kang}}
\author[1]{\fnm{Jeffrey D.} \sur{Hyman}}
\author[3]{\fnm{Abigail} \sur{Hunter}}
\author[4]{\fnm{Nathan} \sur{DeBardeleben}}
\author[5]{\fnm{Earl} \sur{Lawrence}}
\author[1]{\fnm{Hari} \sur{Viswanathan}}
\author[1]{\fnm{Daniel} \sur{O'Malley}}
\author[1]{\fnm{Javier E.} \sur{Santos}}

\affil[1]{\orgname{Energy and Natural Resources Security Group, EES-16, Earth and Environmental Sciences Division, Los Alamos National Laboratory}, \country{USA}}
\affil[2]{\orgname{National Security Earth Science Group, EES-17, Earth and Environmental Sciences Division, Los Alamos National Laboratory}, \country{USA}}

\affil[3]{\orgname{Materials and Physical Data Group, XCP-5, X Computational Physics Division, Los Alamos National Laboratory}, \country{USA}}

\affil[4]{\orgname{High Performance Computing Design Group, HPC-DES, High Performance Computing Division, Los Alamos National Laboratory}, \country{USA}}

\affil[5]{\orgname{Statistical Sciences Group, CCS-6, Computer, Computational and Statistical Sciences Division, Los Alamos National Laboratory}, \country{USA}}
\vspace{-2cm}

\abstract{

Accurately predicting when and how materials fail is critical to the design of safe and reliable structures, mechanical systems, and engineered components that operate under stress. Yet, fracture behavior remains difficult to model across the diversity of materials, geometries, and loading conditions encountered in real-world applications.  While machine learning (ML) methods have shown promise, most models are trained on narrow datasets, lack robustness, and struggle to generalize to new regimes. Meanwhile, physics-based simulators offer high-fidelity predictions but are fragmented across specialized numerical methods and require substantial high-performance computing resources to explore the input space.

To address these limitations, we present a data-driven foundation model for fracture prediction, a transformer-based architecture that operates across simulators, a wide range of materials (including plastic-bonded explosives, steel, aluminum, shale, and tungsten) and diverse loading conditions. The model supports both structured and unstructured meshes, combining them with large language model embeddings of textual input decks that specify material properties, boundary conditions, and solver settings. This multimodal input design enables flexible adaptation across diverse simulation scenarios without requiring changes to the model architecture. The trained model can be fine-tuned with minimal data on diverse downstream tasks. These include time-to-failure estimation, modeling the temporal evolution of fractures, and adapting to combined finite-discrete element method simulations. The model also generalizes to previously unseen materials such as titanium and concrete, requiring as few as a single sample, dramatically reducing data requirements compared to standard ML approaches. Our results demonstrate that fracture prediction can be unified under a single model architecture, offering a scalable, extensible alternative to simulator-specific workflows.
}


\maketitle

Fractures govern the limits of performance and safety in both engineered and natural systems. From structural collapse and brittle failure in critical infrastructure~\cite{qu2024review} to pipeline rupture~\cite{niazi2021high}, the consequences of material failure span a wide range of applications. In contexts such as hydraulic fracturing and subsurface containment of hazardous materials 
~\cite{hyman2016understanding}, understanding when and how geo-materials break is essential not only for mitigating risks such as fluid leakage and induced seismicity, but also for the design and optimization of subsurface engineering systems~\cite{launey2009fracture,kalthoff1986fracture}.

Despite decades of research, reliably predicting fracture behavior across materials and loading conditions remains a fundamental challenge. This has led to the development of numerous specialized numerical simulators, each tailored to address specific modeling hurdles. Mesh-free and particle methods, such as peridynamics and the material point method, handle extreme deformations without mesh entanglement~\cite{silling2000reformulation,sulsky1995application}. Interface-enriched finite elements like XFEM embed discontinuities directly in the solution space, capturing sharp cracks without remeshing~\cite{belytschko1999elastic}. Phase-field models regularize cracks as diffusive fields, allowing for branching and merging on fixed grids~\cite{bourdin2000numerical}, while cohesive-zone and adaptive meshing techniques explicitly evolve mesh topology to follow complex crack paths~\cite{ortiz1999finite}. Although effective within their respective regimes, these methods often suffer from limitations such as high computational cost, the need for a priori knowledge of crack paths, missing material models, and reliance on expert parameter tuning for physically consistent results.
Practitioners often face uncertainty in selecting the appropriate fracture simulator. Each solver family has been developed, benchmarked, and validated within its regime, but lacks interoperability; transferring between them requires re-encoding meshes, constitutive laws, and boundary conditions. We address this gap with a simulation-level foundation model that learns a shared latent representation across diverse fracture solvers. Using inexpensive rule-based surrogate simulations for pretraining, the model learns generic topological operations such as crack growth and coalescence, then adapts to high-fidelity phase-field and finite-discrete element data via a multi-fidelity curriculum. This strategy reduces data demands and enables rapid cross-solver inference for example, replacing XFEM screening sweeps with single-GPU predictions. Similar multi-fidelity approaches are already used in aerospace, where Gaussian-process surrogates and neural networks combine RANS and LES data for efficient aerodynamic optimization~\cite{scoggins2023multihierarchy, tao2024multifidelity}.

Recently, machine learning (ML) approaches have emerged as a tool that offers the potential to learn predictive models directly from data. Among these, foundation models~\cite{bommasani2021opportunities} are large machine learning models trained on broad and diverse datasets, enabling them to learn general-purpose representations that can be adapted to a wide range of downstream tasks. These models are typically pretrained and fine-tuned for specific applications, leveraging their ability to transfer knowledge across domains. Their defining features include scalability, versatility, and strong performance with minimal task-specific data. Foundation models provide a framework for capturing patterns across domains and efficiently adapting to new tasks by exploiting knowledge acquired during pretraining~\cite{brown2020language}. Applications in weather forecasting~\cite{bodnar2025aurora,palmer2008toward}, protein structure prediction~\cite{jumper2021highly,stokes2020deep}, and learning interatomic potentials for molecular dynamics~\cite{rupp2012fast,ahmad2022chemberta,batatia2023foundation} demonstrate that training on diverse datasets can yield high-performing systems that generalize across tasks and offer a scalable, efficient alternative to traditional simulation or data collection pipelines by aggregating available information and delivering predictions orders of magnitude faster than conventional methods.
Foundation models are increasingly impacting materials science~\cite{pyzer2025foundation,pyzer2022accelerating}: enabling automatic data extraction from the literature~\cite{gupta2022matscibert,rajan2023decimer}, prediction of molecular properties and structures~\cite{takeda2023foundation,chithrananda2020chemberta,ross2022large,bagal2021molgpt}, and support for material synthesis~\cite{chang2024bidirectional,kim2020inorganic}. 

Specifically for fracture propagation in materials, ML methods have been applied to predict crack evolution~\cite{lew2021deep, perera2022graph, hsu2020using, zhang2022prediction, perera2023dynamic, chen2024hossnet}, crack coalescence~\cite{moore2018predictive, hunter2019reduced}, stress distribution~\cite{wang2021stressnet, perera2023dynamic, perera2022graph}, fracture toughness~\cite{liu2020machine, daghigh2020machine}, fatigue strength~\cite{yan2020predictions, frie2024exploration, wang2024recent}, time to failure for materials~\cite{hunter2019reduced, biswas2020prediction}, and accelerating workflows~\cite{liu2022generalized, perry2022automated, thi2025concrete}. However, most ML models for fracture propagation are typically trained on narrow datasets from a single simulator, under fixed material and geometric conditions. As a result, they struggle to generalize across simulation settings, limiting their practical utility and reproducibility.

Transfer learning has been shown to alleviate some of these concerns~\cite{yosinski2014transferable, zhou2024transfer, perera2023generalized}, but the required training data and necessary modifications to the network architecture  can vary significantly depending on the problem statement~\cite{zhao2024comparison}. Alternatively, the rise of large-scale foundation models trained on heterogeneous datasets offers a path towards more general models that retain accuracy across data types and tasks, enabling the development of material discovery workflows~\cite{dijkstra2021predictive,ekins2019exploiting}

We present the first foundation model tailored to material failure prediction, introducing a novel multimodal architecture that unifies structured simulation data and textual metadata within a single predictive framework. Unlike prior work, which relies on simulator-specific models trained on narrow datasets, our approach combines spatial fracture fields with large language model (LLM) embeddings of input decks, enabling the model to interpret material properties, boundary conditions, and solver configurations directly from text. This design allows us to generalize across fracture simulation regimes without hand-engineered pipelines or architectural changes. Our two-stage training strategy: pretraining on a multi-fidelity dataset that includes both rule-based surrogates and high-fidelity physics-based simulations, followed by task-specific fine-tuning. The model achieves high accuracy at a fraction of the computational cost of traditional numerical solvers, while remaining adaptable across materials, loading conditions, and simulation frameworks. Specifically the foundational model has the following key features.

\begin{itemize}
\item   \emph{Multifidelity training across simulators}: 
        Our approach enables scalable pretraining in regimes where high-fidelity physical simulations are limited in availability. We develop and use a rule-based surrogate simulator capable of generating new training simulations on-the-fly. We demonstrate that models trained with a combination of surrogate and limited high-fidelity simulation data can match the performance of models trained on one order of magnitude more simulation data, substantially reducing data requirements. 


\item   \emph{Generalization across initial fracture patterns}: 
        The foundational model generalizes to unseen fracture patterns (including variations in crack density, orientation, and spatial configuration) despite being trained on a narrow subset of simulation types. This capability is enabled by the transformer's global receptive field, which allows it to capture long-range spatial dependencies across the domain.

\item   \emph{Generalization to new materials}: 
        New materials can be incorporated through minimal fine-tuning (requiring as few as a single example) without retraining the full model. This approach supports future extensibility as new materials and simulation types become available.

\item   \emph{Predicting new quantities of interest}: 
        The foundational model supports additional prediction tasks, such as time-to-failure estimation, achieving less than 10\% mean relative error with decoder-only fine-tuning and minimal compute overhead.

\item   \emph {Temporal evolution predictions}:
        Although trained on static images, the model can be fine-tuned to predict the full temporal evolution of the failure process. While training used ten discrete simulation steps, we show that the model can generate intermediate states at inference time, enabling continuous interpolation of the fracture trajectory.

\item   \emph{Generalization across mesh types}:
        Although the model was exclusively trained on Cartesian grids, it can be fine-tuned to unstructured meshes with high accuracy. We demonstrate this on high-fidelity simulations from Hybrid Optimization Software Suite (HOSS)~\cite{Knight_etal_2020}, a computationally intensive combined finite-discrete element method~\cite{munjiza_combined_2004, munjiza_large_2015, munjiza_compmechdis_2011} implementation, achieving fracture predictions in a fraction of the simulation time.
        
\end{itemize}

\section{A Foundation Model for Material Failure Prediction}

\begin{figure}
    \centering
    \includegraphics[width=\textwidth]{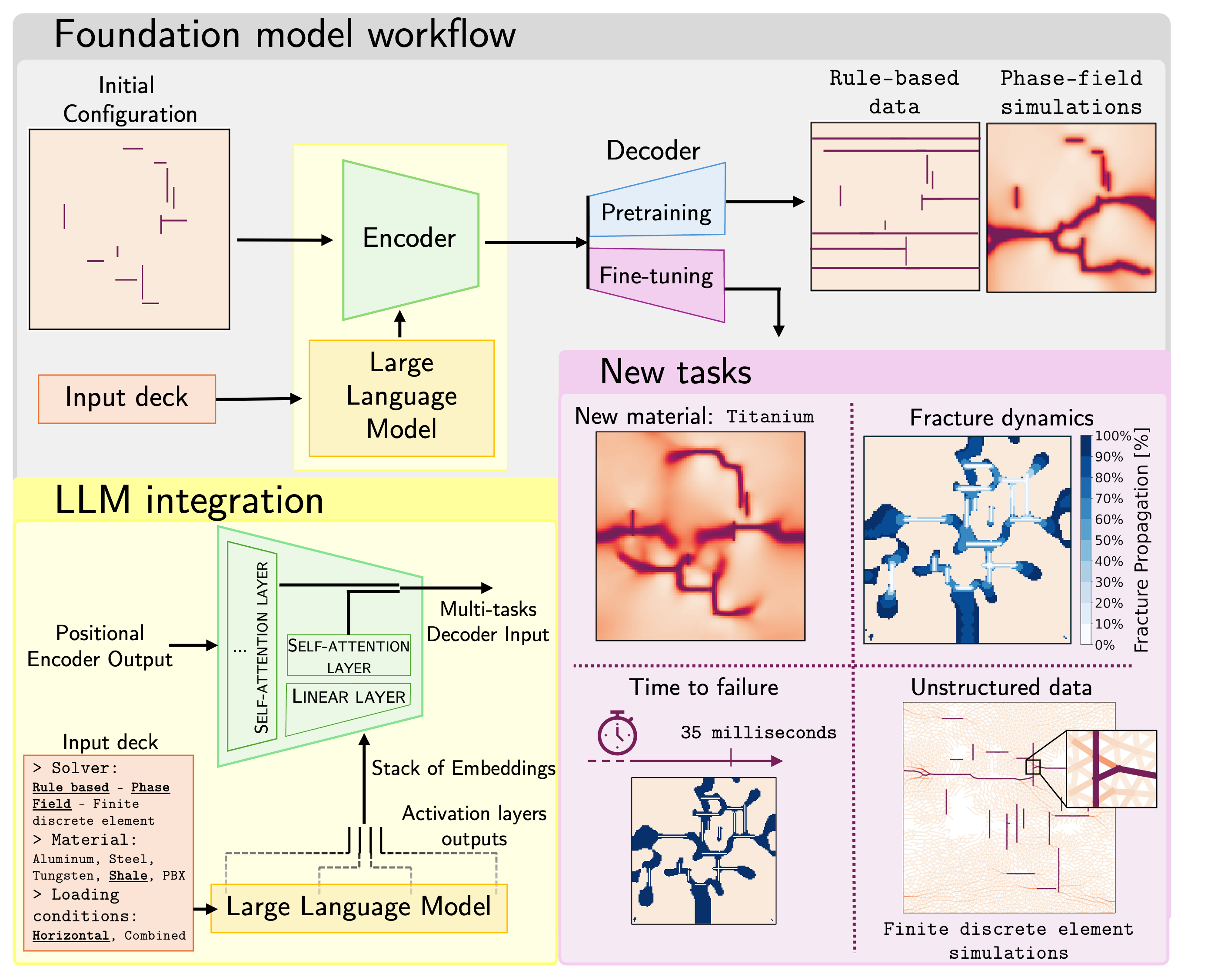}
    \caption{Our foundation model demonstrates the ability to predict failure behavior of unseen materials with as few as a single sample. (\textbf{Grey}) Foundation model workflow: an encoder-decoder transformer is pretrained on rule-based data and phase field simulations to predict failure patterns from initial configurations of materials under axial or biaxial pulling. (\textbf{Yellow}) LLM integration: an input deck containing information about the material, simulator, and loading conditions is processed via latent representations from intermediate layers of a large language model. (\textbf{Purple}) New tasks: the pretrained model is fine-tuned to tackle downstream tasks, including fracture dynamics, time-to-failure prediction, failure pattern inference from unstructured simulators, and generalization to new materials.}
    \label{fig:overview}
\end{figure}

We introduce a foundation model designed to predict fracture behavior across a broad spectrum of materials (metals, geomaterials, and high-energy composites), as well as diverse geometries, and simulation settings. Through fine-tuning, the model supports multiple tasks, including fracture pattern prediction, time-to-failure estimation, and dynamic evolution of fractures on both Cartesian and unstructured grids.

The architecture is composed of three components: (1) an encoder that transforms spatially structured or unstructured input grids into a unified latent representation (Figure~\ref{fig:overview}A); (2) a textual input deck context encoder that integrates auxiliary simulation metadata using large language model (LLM) embeddings (Figure~\ref{fig:overview}B); and (3) a task-specific decoder that generates outputs corresponding to the selected prediction task (Figure~\ref{fig:overview}C).

Contextual metadata (such as material properties, loading conditions, and simulator identifiers) is processed through a LLM pipeline (LLaMA-3.1 8B~\cite{grattafiori2024llama}) and fused with the spatial latent representation. This design enables the model to adapt to new simulation settings without structural changes and facilitates 
generalization to unseen configurations. It also allows external users to fine-tune the model on their own data with minimal effort, without retraining the full network.

To train the model, we leverage a diverse dataset spanning synthetic and 
physics-based simulations. Pretraining is performed on a mixture of low-cost surrogate simulations and 
physics solvers, allowing the model to learn general-purpose representations of fracture behavior across different materials, loading conditions, and numerical fidelity. These representations encode both physical behavior and computational representations, such as mesh geometry and boundary conditions, which influence fracture dynamics. The base model was pretrained using phase-field-based fracturing simulations \cite{Hirshikesh_etal_2019,Zhou_Zhuang_2020} of different materials (plastic-bonded explosives, steel, aluminum, shale, and tungsten) and loading conditions (axial and biaxial pulling). 

Pretraining was conducted over seven million optimization steps on 64 NVIDIA H100 GPUs, over approximately eight days using data parallelism. This hardware configuration serves as the reference for all training times reported in the paper. The model was trained to minimize the mean squared error (MSE) between predicted and simulated final fracture patterns, using a combination of rule-based and phase-field simulations. Details about the generation of the data can be found in Section~\ref{sec:data} of the supporting information.

Test errors are evaluated using the L1 loss, a robust and interpretable metric for image-to-image prediction tasks across heterogeneous datasets and tasks:
\begin{equation}
    L_1 = \frac{1}{N} \sum_{i=1}^{N} |y_i - \hat{y}_i|
\end{equation}
\noindent where \( N \) is the total number of elements in the simulation, \( y_i \) is the ground truth value, and \( \hat{y}_i \) is the predicted value. 

\section{Pretraining}

\begin{figure}
    \centering
    \includegraphics[width=0.8\textwidth]{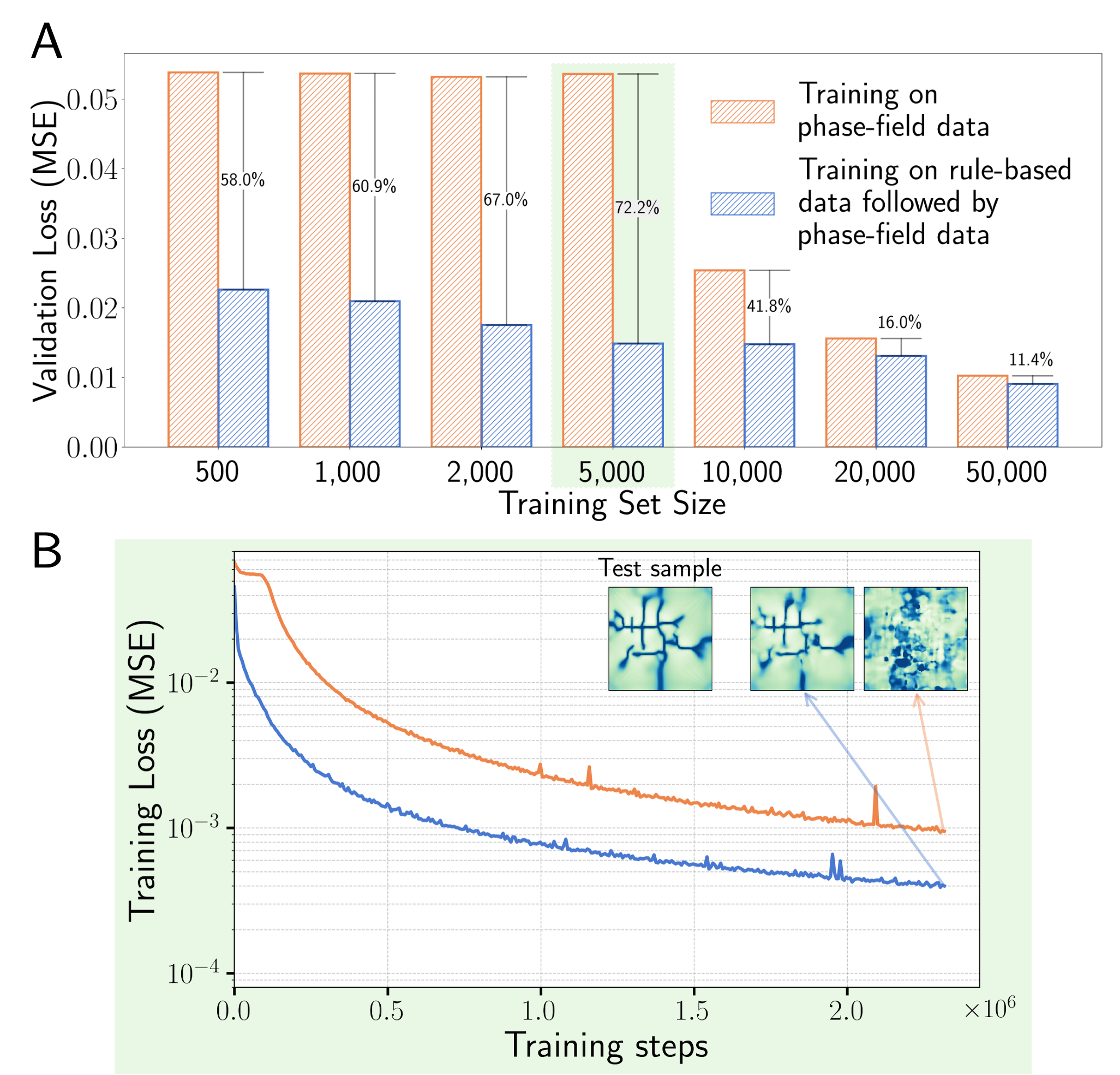}
    \caption{Pretraining on inexpensive rule-based data can reduce errors by more than 70\% for failure pattern prediction. (A) Minimum validation loss as a function of the training set size. Models trained solely on phase field data (orange) are compared with models trained with rule-based data and the same phase-field simulations (blue). Pretraining improves performance in low-data regimes. (B) Training loss over time for a dataset with 5,000 samples. Insets show predictions of the failure pattern at different training stages for a test sample, highlighting faster convergence and improved accuracy with pretraining.}
    \label{fig:pre-pre-training}
\end{figure}

\begin{figure}
    \centering
    \includegraphics[width=\textwidth]{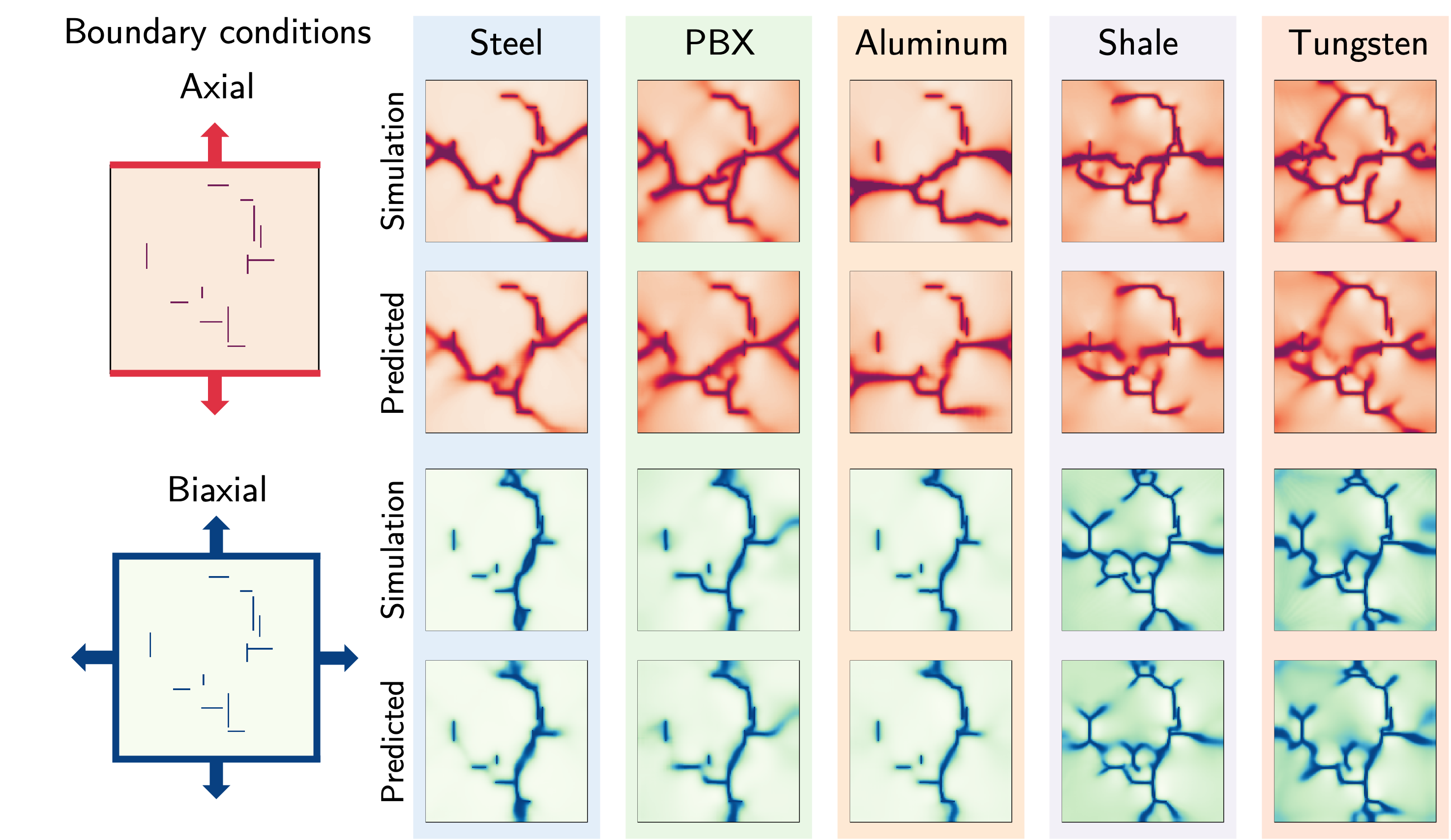}
    \caption{Our foundation model generalizes across materials (steel, PBX, aluminum, shale, and tungsten) and loading conditions: axial loading (top rows) and biaxial loading (bottom rows). All samples share the same initial configuration (left). Colorbars are normalized from 0 to 1.
    }
    \label{fig:kai_diff_BCs}
\end{figure}

Pretraining on simplified data has proven remarkably effective across scientific and engineering domains, providing a strong prior for downstream tasks with limited supervision.  For example, Aurora, a foundation model for weather forecasting~\cite{bodnar2025aurora},  achieves state-of-the-art accuracy by first pretraining on raw, uncurated simulation outputs that capture physical dynamics without manual labeling or curating specialized satellite data. Similarly, V-JEPA 2~\cite{assran2025vjepa2} learns powerful visual representations through  pretraining on over a million hours of unlabeled internet videos, while large language models (such as GPT-4~\cite{openai2024gpt4} and LLaMA~\cite{touvron2023llama}) are pretrained on massive, unfiltered text corpora to acquire general-purpose representations that are later refined via lightweight task-specific tuning.


Inspired by these ideas, we implemented a simple rule-based algorithm as a surrogate model to simulate fracture propagation in a Cartesian lattice. In this algorithm, initial horizontal and vertical fractures are placed in the domain and expand toward the boundaries according to distinct rules (analogous to subsurface processes as detailed in Section~\ref{sec:fwb} of the supporting information). This data is inexpensive to generate, in fact, can be produced on the fly during training, enabling an effectively endless stream of diverse configurations. This surrogate pretraining phase enables the model to learn basic geometric and topological patterns of fracture propagation, while establishing associations between the encoded input decks and the resulting fracture fields at minimal computational cost.

We investigated how training on surrogate rule-based data influences the model's ability to predict fracture behavior in phase-field simulations - the target data we surrogate with this foundation model. Training first with rule-based data improves the expressiveness of the learned representations, particularly when high-fidelity data are limited (due to cost and/or runtime). In Figure~\ref{fig:pre-pre-training}A we compare a model trained solely on phase-field simulations with one that is first trained on rule-based data and subsequently on the same set of phase-field examples. The latter achieves a 72\% improvement in validation performance with only 5,000 training samples. Figure~\ref{fig:pre-pre-training}B illustrates that the model trained with limited phase-field data struggles to learn coherent spatial patterns of fracture propagation (likely due to the weak inductive biases of transformers) whereas the model exposed to rule-based captures more robust and meaningful features. As the size of the fine-tuning dataset increases, the performance gap narrows, but models trained with rule-based data continue to converge faster and reach lower final loss. We attribute this to the transformer's limited structural priors: in the absence of architectural constraints such as convolutions, exposure to structured training data provides  guidance toward physically meaningful solutions. The complete training and validation loss curves are shown and described in Figure~\ref{fig:training-set-loss} in the supporting information.

\begin{figure}
    \centering
    \includegraphics[width=\textwidth]{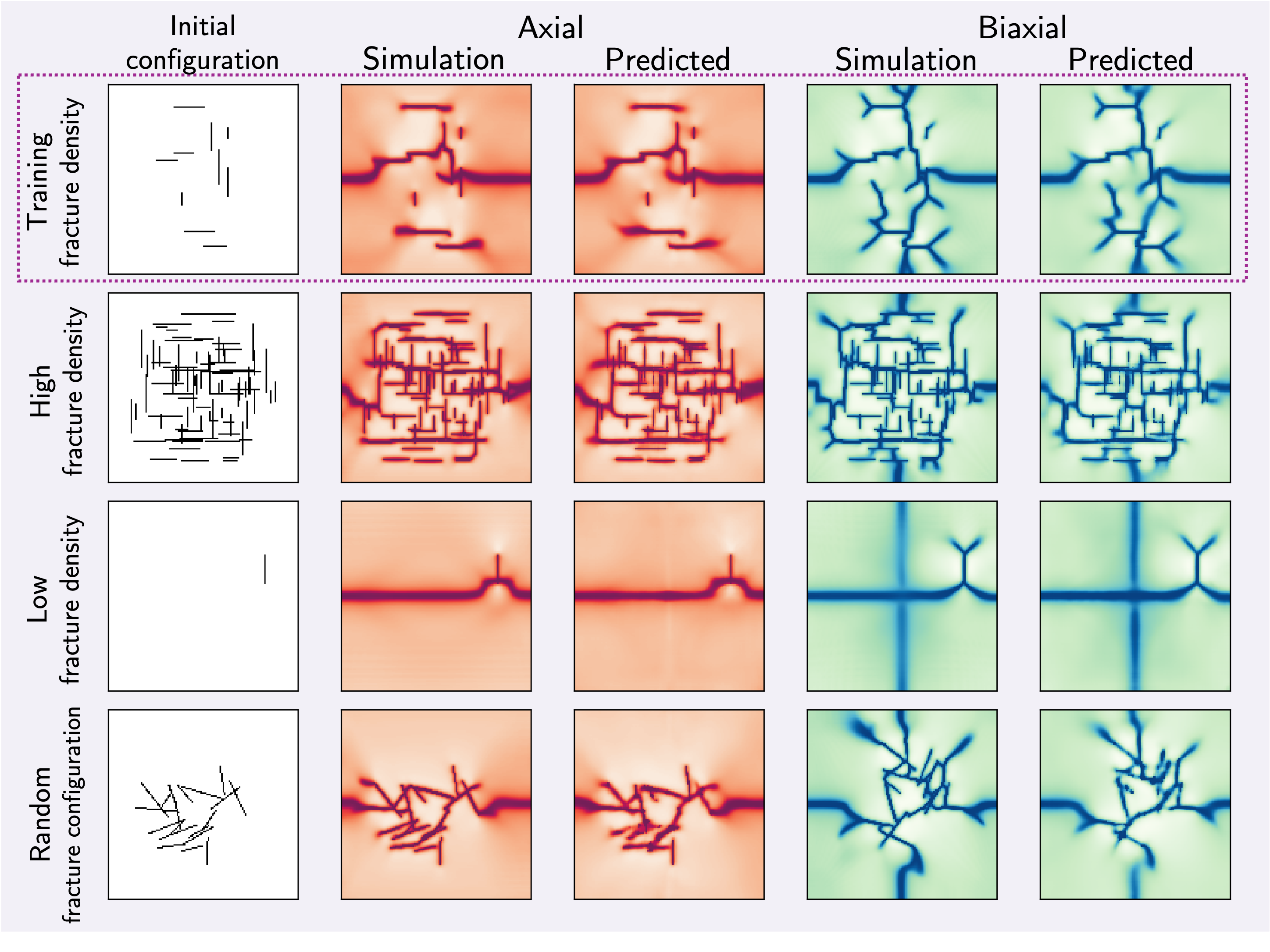}
    \caption{Our foundation model generalizes to in- and out-of-distribution (i.e., not seen during training) initial fracture configurations for shale material. Each row represents a distinct type: training-like (top), high fracture density, single fracture, and random orientations. The columns represent the initial configuration of the fractures in the domain, the simulated and predicted failure patterns (for axial and biaxial boundary conditions).}
    \label{fig:kai_generalization}
\end{figure}

Model accuracy improved consistently with transformer size, for the experiments reported in this work, we selected the 20M parameter configuration to balance predictive performance with computational efficiency. A detailed scalability study is reported in Figure~\ref{fig:scalability} of the supporting information.

To evaluate performance at scale, we trained the model on the full dataset of 2 million phase-field simulations. This resulted in strong generalization capabilities, as shown in Figure~\ref{fig:kai_diff_BCs}, where the model accurately predicts failure behavior across boundary conditions and material types on held-out test data, achieving a validation loss of 0.003.

Although the model is pretrained on a narrow set of initial fracture configurations (from 6 to 30 horizontal and vertical fractures), it generalizes accurately to out-of-distribution cases. These include domains with a single initial fracture, high-density fracture fields, and oblique or diagonal orientations not seen during training. 
A qualitative comparison of a selection  of these cases are presented in Figure~\ref{fig:kai_generalization}. We attribute this robustness to the global receptive field of the transformer architecture, which enables non-local reasoning across the entire spatial domain. Unlike convolutional models, which rely on local kernels, our approach captures long-range dependencies critical for modeling fracture interactions and propagation. A larger collection of predictions is provided in Section~\ref{sec:add_pred} of the supporting information. The mean absolute errors on the test set for the different materials, boundary conditions, and generalization case are reported in Table~\ref{tab:mae_results_generalization} of the supporting information.


\section{Fine-tuning}

As shown in the ``New tasks" panel of Figure~\ref{fig:overview}, the pretrained model can be fine-tuned for a range of downstream tasks, including fracture propagation dynamics, time-to-failure estimation, and generalization to new materials or mesh types. The fine-tuning tasks do not require changes in the architecture, so the knowledge acquired in the pretraining phase is exploited in the adaptation to the new task. The fine-tuning strategy, which includes the dataset size and the part of the architecture being trained, depends on the specific task, as detailed in the following sections.

\subsection{New materials}

For this fine-tuning task, the objective is to train the model to predict failure patterns for new materials not present in the base model's training dataset, specifically, titanium and concrete. The flexibility of the LLM-based input deck encoding allows us to introduce new materials by simply specifying their names. In order to reuse the representation learned during the pre-training phase,  we trained only the decoder's parameters (30\% of the total number of training parameters) for 1,500 steps, with this strategy the training takes approximately three minutes. 

\begin{figure}
    \centering
    \includegraphics[width=0.75\textwidth]{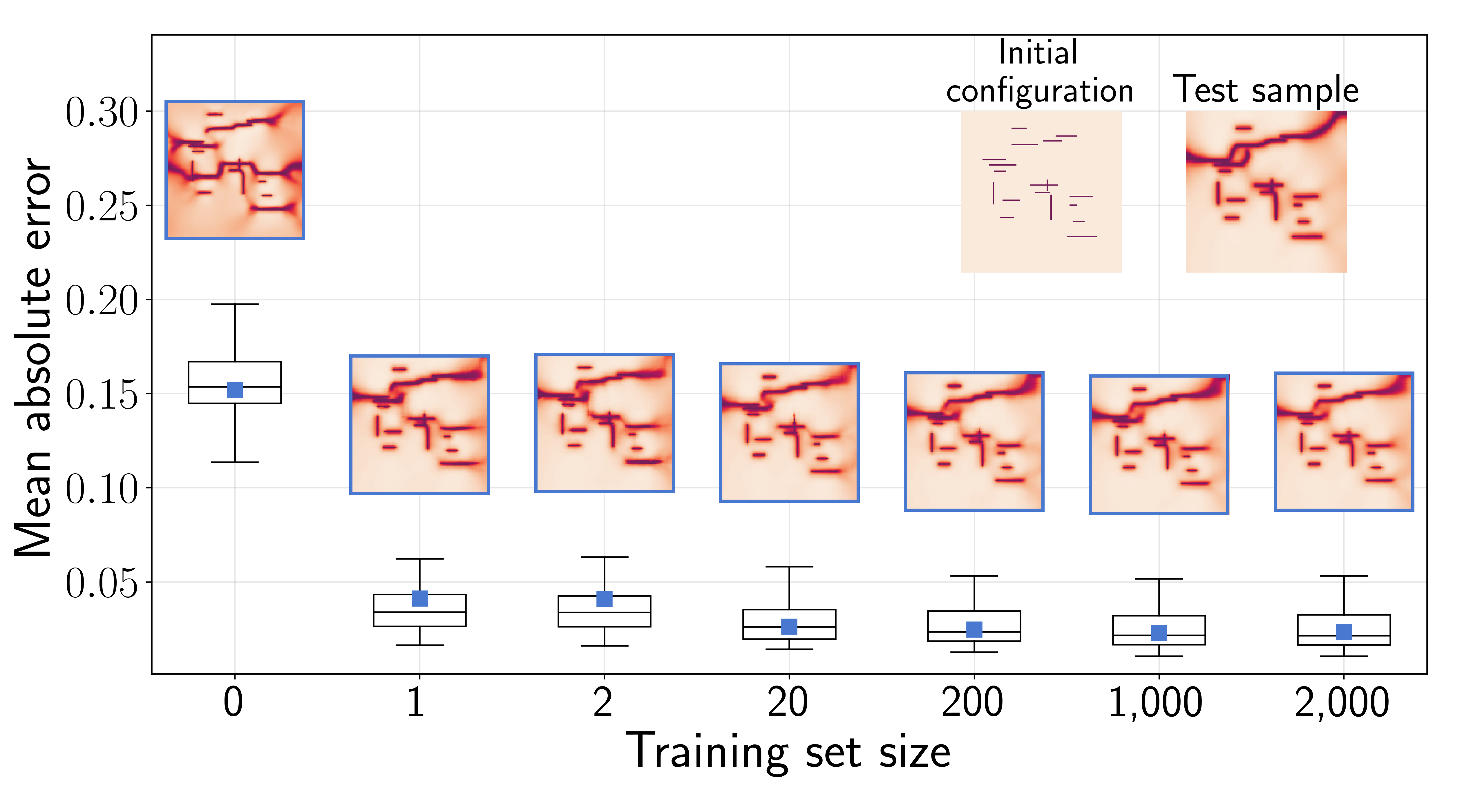}
    \caption{Finetuning on one sample enables accurate predictions for titanium, a material not included in the base model pretraining. The distributions of mean absolute error across test samples for each training set size are shown. For each case, a representative predicted failure pattern is shown above its corresponding error field (difference from ground truth), with the ground truth and initial configuration displayed in the top right corner. The blue square marker in each boxplot indicates the mean error associated with the selected representative prediction.}
    \label{fig:fine-tuning-new-material}
\end{figure}

We evaluated the performance of the model as a function of the number of training samples, as shown in Figure~\ref{fig:fine-tuning-new-material}. As expected, we observe a decrease in the MAE with increasing training set size for titanium simulations. However, the errors stabilized after 1 sample, indicating no noticeable improvements with further increase in training dataset. This may be due in part to overlap in failure behavior with materials seen during pretraining. The number of samples required for effective fine-tuning varies by material. In contrast, concrete requires more training samples due to its more distinct failure behavior. More details about the generalization to concrete can be found in Section~\ref{sec:concrete} of the supporting information.

%


\subsection{Temporal prediction}

While the final fracture state is crucial for assessing material failure, understanding how fractures evolve from initial defects offers valuable insight into the underlying failure dynamics. To model this evolution, we fine-tuned the decoder to predict intermediate states along the fracture trajectory. We exploited the flexibility of the LLM embeddings to introduce an additional input: the progression point, specified as a normalized scalar ranging from 0.3 (start of the fracture propagation in the phase-field simulations) to 1 (fully developed fracture).

\begin{figure}
    \centering
    \includegraphics[width=0.7\textwidth]{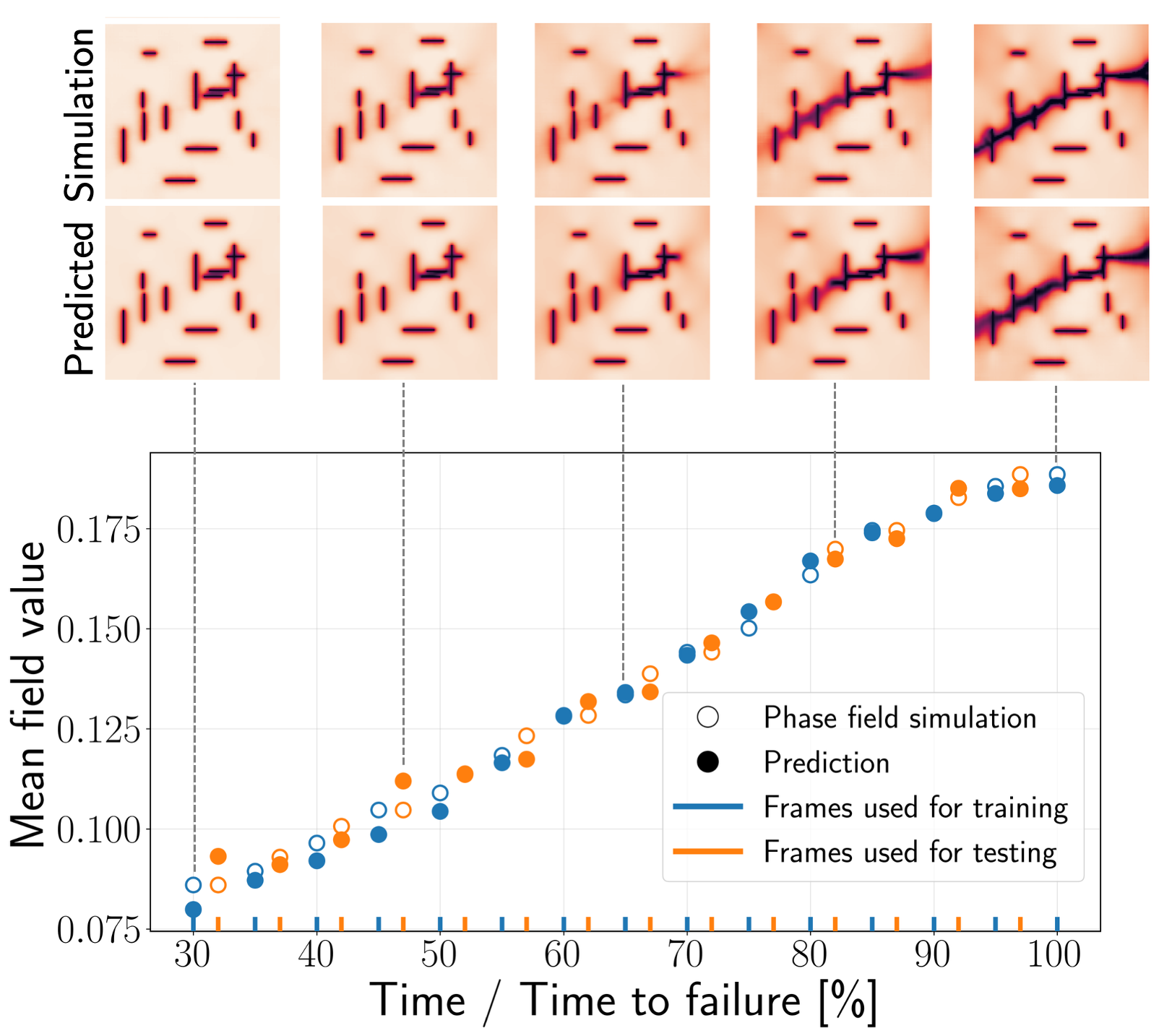}
    \caption{The foundation model, which is pretrained to predict final fracture patterns, can be finetuned to learn fracture dynamics: snapshots of fracture propagation (top) and evolution of the mean field value (bottom) with training frames (blue) and interpolated predictions (red). }
    \label{fig:fine-tuning-temporal}
\end{figure}

We trained the decoder to evolve the initial state to the specified point in the progression. We trained it using ten uniformly spaced steps along the trajectory. In Figure~\ref{fig:fine-tuning-temporal} we show the dynamics prediction on a held-out sample, the model produces highly accurate predictions across the entire range of training progression points (blue). Notably, it also generalizes to unseen intermediate values (orange), for instance, generating coherent and physically plausible outputs at 0.63 progression, despite never being explicitly trained on that progression point. This demonstrates the effectiveness of our LLM embeddings in conveying contextual information to the network. The training time for this task is 16 hours.


\subsection{Time to failure}

\begin{figure}
    \centering
    \includegraphics[width=0.6\textwidth]{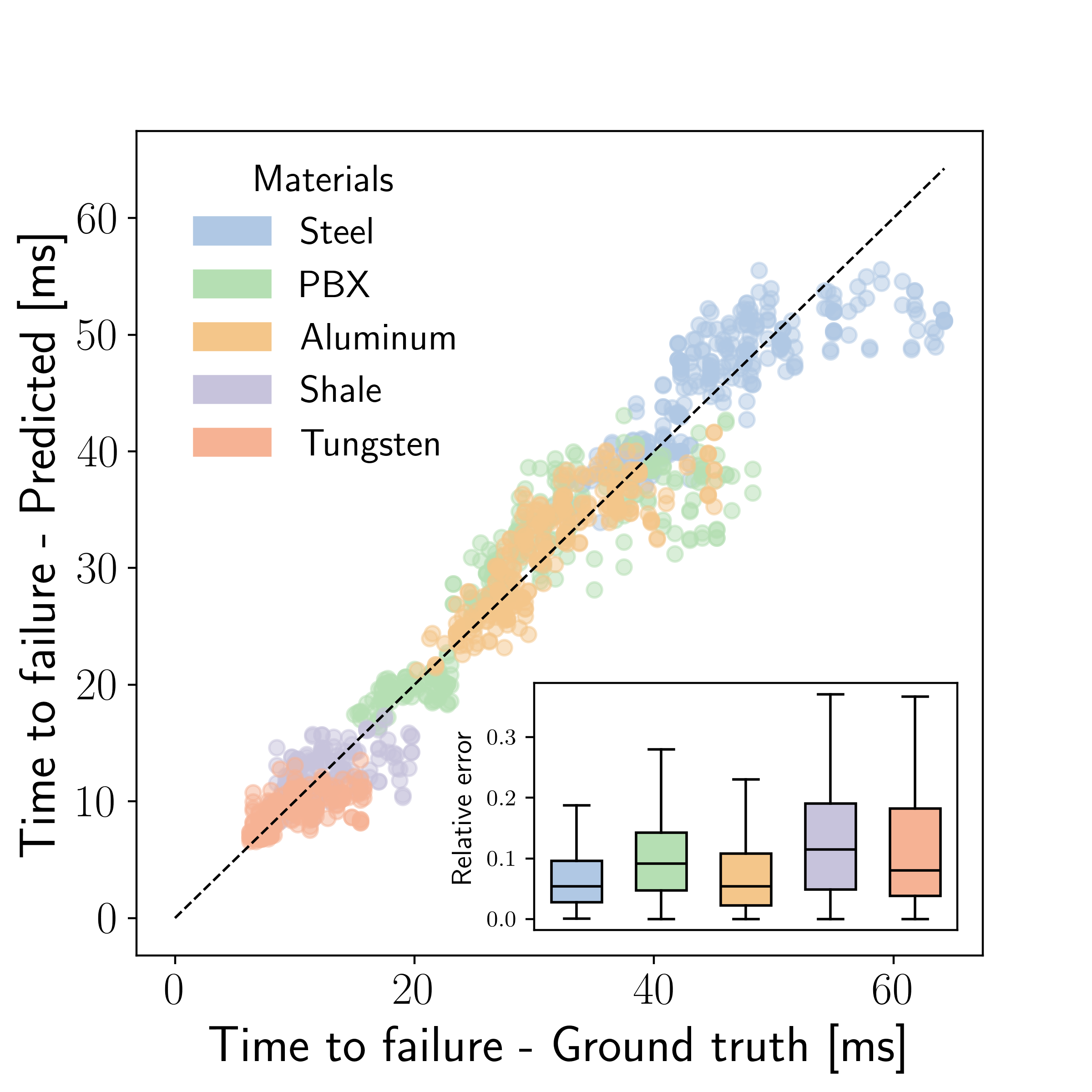}
    \caption{With finetuning, the model generalizes beyond prediction fracture patterns and is able to predict the time to failure. The plot shows predicted vs. ground truth time to failure on the test set of phase-field simulations. Inset shows the distribution of relative errors across materials. }
    \label{fig:fine-tuning-ttf}
\end{figure}

In many applications, the time to failure alone is the key quantity of interest~\cite{liu2024microstructure, guarino2002failure,hassine2014time}, regardless of how the fracture evolves or the way it breaks spatially. To predict this scalar value, we fine-tuned the model to estimate time-to-failure from the initial defect configuration and material context. Only the decoder was trained for this task, with an added postprocessing layer to map the output to a single scalar. The model achieves strong predictive performance, reaching an $R^2$ of 0.9522 across materials and loading conditions in held-out simulations, outperforming previous works with a performance increase of the 30\%~\cite{moore2018predictive}. The distribution of the time to failure prediction error is shown in Figure~\ref{fig:fine-tuning-ttf}. Training required approximately 16 hours.

\subsection{Capturing fracture dynamics with unstructured grids}

\begin{figure}
    \centering
    \includegraphics[width=\textwidth]{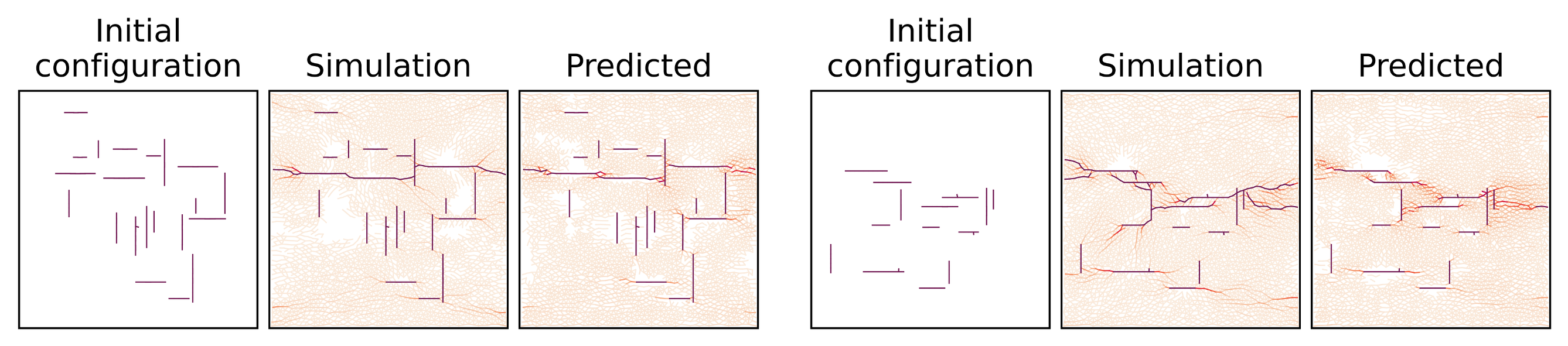}
    \caption{Generalization to two test samples from a finite discrete element simulation with an unstructured mesh based on HOSS (FDEM) simulations that can handle large deformations.}
    \label{fig:fine-tuning-hoss}
\end{figure}

The architecture is agnostic to mesh structure, resolution, and dimensionality. To assess this capability, we fine-tuned the model on finite-discrete element method simulations generated using the Hybrid Optimization Software Suite (HOSS), a high-fidelity solver that captures detailed fracture dynamics on complex unstructured grids~\cite{Knight_etal_2020}, more details about the data generated can be found in Section~\ref{sec:hoss-pbx} of the supporting information. In this case, the different coordinate structure required training the full architecture rather than just the decoder. Despite the topological variability and geometric complexity of the HOSS data, the model adapted effectively and produced accurate predictions for PBX. Training completed in approximately 16 hours, yielding a MAE of 0.0218, examples of predictions are shown in Figure~\ref{fig:fine-tuning-hoss}.

\section{Discussion}

We presented a foundation model for material failure prediction, capable of generalizing across simulators, materials, grid types, and tasks. By leveraging a lightweight surrogate model during pretraining, contextual textual input deck encodings, and a flexible transformer-based architecture, the model supports rapid fine-tuning across a diverse range of downstream applications including fracture progression, time-to-failure estimation, and adaptation to unstructured meshes. Fine-tuning on each of these tasks required only minutes to hours, in contrast to the significantly longer development cycles typical of conventional machine learning pipelines. It is important to emphasize, however, that this accelerated workflow is enabled by the extensive availability of high-quality simulation data, which stems from decades of in-house research and investment in traditional numerical solvers.

Scaling pretraining to broader datasets and larger model sizes may further improve both accuracy and generalization, particularly when incorporating a wider range of physical phenomena. For example, the model could be fine-tuned to predict fracture behavior directly from image-based observations of initial conditions, enabling rapid analysis of physical samples in experimental settings. A current limitation is that the model and training data is composed solely of fracture evolution under extensional loading. We would expect increased mode~II crack propagation and markedly different fracture emergence and evolution under  shear-dominated or compressional loading. While this work focused on two-dimensional simulations due to the prohibitive cost of high-resolution 3D fracture modeling, the required architectural modifications are minimal.  Extending to 3D or to more complex two-dimensional geometries is technically straightforward within our framework.
Finally, interpretability also remains an open direction. Identifying which internal representations align with known physical mechanisms could offer valuable insights into both model reasoning and material response, and may ultimately help build trust in these models for scientific and engineering applications.

Our results suggest that fracture prediction, a historically fragmented and simulation-specific task, can be unified under a single general-purpose model.  More broadly, the multimodal architecture introduced here, combining spatial simulation outputs with LLM-encoded contextual metadata, may be applicable to a wide range of problems in computational science, where simulations are governed by both physical fields and structured parameter inputs. We see this as a promising step toward general-purpose surrogate models that interface directly with existing simulation pipelines, accelerating scientific workflows across domains.


\section*{Methods}

\subsection*{Foundation model architecture}

The model architecture adopts an encoder–decoder transformer design tailored for scientific simulations, with the goal of unifying diverse fracture prediction tasks under a single framework (Figure~\ref{fig:overview}). Although encoder–decoder models are common in machine learning, their application to spatially heterogeneous scientific data, such as structured and unstructured meshes, temporal trajectories, and multimodal physical outputs, remains an open research area.

In contrast to prior work, our architecture integrates topological flexibility (handling both Cartesian and unstructured grids) with task diversity (predicting fracture fields, time-to-failure, and temporal evolution) in a single transformer backbone. This design imparts strong structural priors, enabling generalization across simulators, material systems, and mesh discretizations.

The architecture comprises three principal components summarized below.

\subsubsection*{Encoder: Flexible input size}

 Let $\mathbf{x} = \{x_1, x_2, \dots, x_N\}$ denote the input tokens derived from a spatial mesh, and $\mathbf{z} = \{z_1, z_2, \dots, z_M\}$ the set of learnable latent tokens. The encoder uses cross-attention to produce a latent representation:
\[
\mathbf{z}' = \text{CrossAttn}(\mathbf{z}, \mathbf{x}) = \text{softmax}\left( \frac{Q_z K_x^\top}{\sqrt{d}} \right) V_x,
\]
where
\[
Q_z = \mathbf{z} W_Q, \quad K_x = \mathbf{x} W_K, \quad V_x = \mathbf{x} W_V.
\]

This yields a size-invariant encoding of spatial inputs, adaptable to arbitrary resolution and topology. All spatial fields are flattened into 1D sequences, allowing the model to operate on domains of arbitrary resolution and topology. Positional encodings anchor the inputs to physical space, enabling seamless use on both structured and unstructured meshes.

Each input element is augmented with a sine–cosine positional encoding of its spatial coordinates, concatenated with the initial fracture configuration to provide geometric context. On structured grids, these positions correspond to Cartesian bin centers, while on unstructured meshes (such as those from HOSS simulations) they align with edge centers where damage occurs. Unlike convolutional neural networks, which require structured grids and cannot natively operate on unstructured meshes without interpolation or remeshing, our transformer-based encoder handles both directly via attention. These are cross-attended with a fixed set of learnable latent tokens that captures the model’s internal state. These tokens interact with the input through the attention mechanism, yielding a compact, size-invariant encoding of the spatial field. To further refine this representation, the latent tokens undergo stacked self-attention layers, enabling long-range reasoning over the geometry and producing a condensed form suitable for spatial prediction and downstream queries.

\subsubsection*{Textual Input Deck creation}

To incorporate simulation metadata, we use a pretrained large language model to generate rich, semantically structured embeddings. Unlike conventional categorical encodings, this approach leverages the semantically-rich compositional priors of modern LLMs to capture complex simulation conditions and helps to model complex relationships among simulation attributes.

Input metadata such as simulation type, material, boundary conditions, and prediction targets are templated as a  natural language prompt. For instance:

\vspace{0.5em}
\noindent \texttt{A \textcolor{gray}{\{simulation\}} of \textcolor{gray}{\{material\}} under \textcolor{gray}{\{boundary conditions\}} to predict \textcolor{gray}{\{target\}}.}
\vspace{0.5em}

\noindent An example instance might read:

\vspace{0.5em}
\noindent \texttt{A simulation using the \textit{\textcolor{gray}{phase-field method}} on \textit{\textcolor{gray}{steel}} under \textit{\textcolor{gray}{axial loading}}, aiming to predict the \textit{\textcolor{gray}{final fracture pattern}}.}
\vspace{0.5em}

Each field is populated from a controlled vocabulary:
\begin{itemize}
\item \textbf{Simulation:} rule-based, phase-field, discrete–finite-element
\item \textbf{Material:} steel, aluminum, PBX, tungsten, shale, concrete
\item \textbf{Boundary conditions:} axial loading, biaxial loading
\item \textbf{Target:} fracture pattern, time to failure, dynamic trajectory
\end{itemize}

These textual descriptions are embedded using LLaMA 3.1 (8B), and their internal representations are fused with the spatial latent tokens via cross-attention. The resulting context vectors allow the model to understand dependencies across simulation regimes, materials, and objectives.

This design enables the model to quickly adapt to new configurations, transfer knowledge across simulation families, and fine-tune with minimal data. For instance, as simulation protocols evolve (runs with different convergence criteria, etc.), updated descriptions can be injected without retraining the full architecture, ensuring future extensibility.

\subsubsection*{Decoder: Multiple downstream tasks}

The decoder translates the encoder's latent representation into task-specific outputs. It operates via cross-attention, conditioning on the encoded spatial and contextual state to support a variety of downstream tasks. The architecture of the decoder accommodates both high-dimensional spatial fields and low-dimensional scalars. For fracture prediction, the decoder produces output fields aligned with the input resolution. For scalar tasks, such as time-to-failure estimation, the representation is compressed to a single floating point number. This unified interface enables the model to flexibly output images, trajectories, or physical quantities, without requiring substantial architectural changes.

\clearpage

\supportinginfo

\begin{center}
    \LARGE \textbf{Supporting Information}
\end{center}

\section{Model size selection}
\label{app:model_selection}

\begin{figure}
    \centering
    \includegraphics[width=\textwidth]{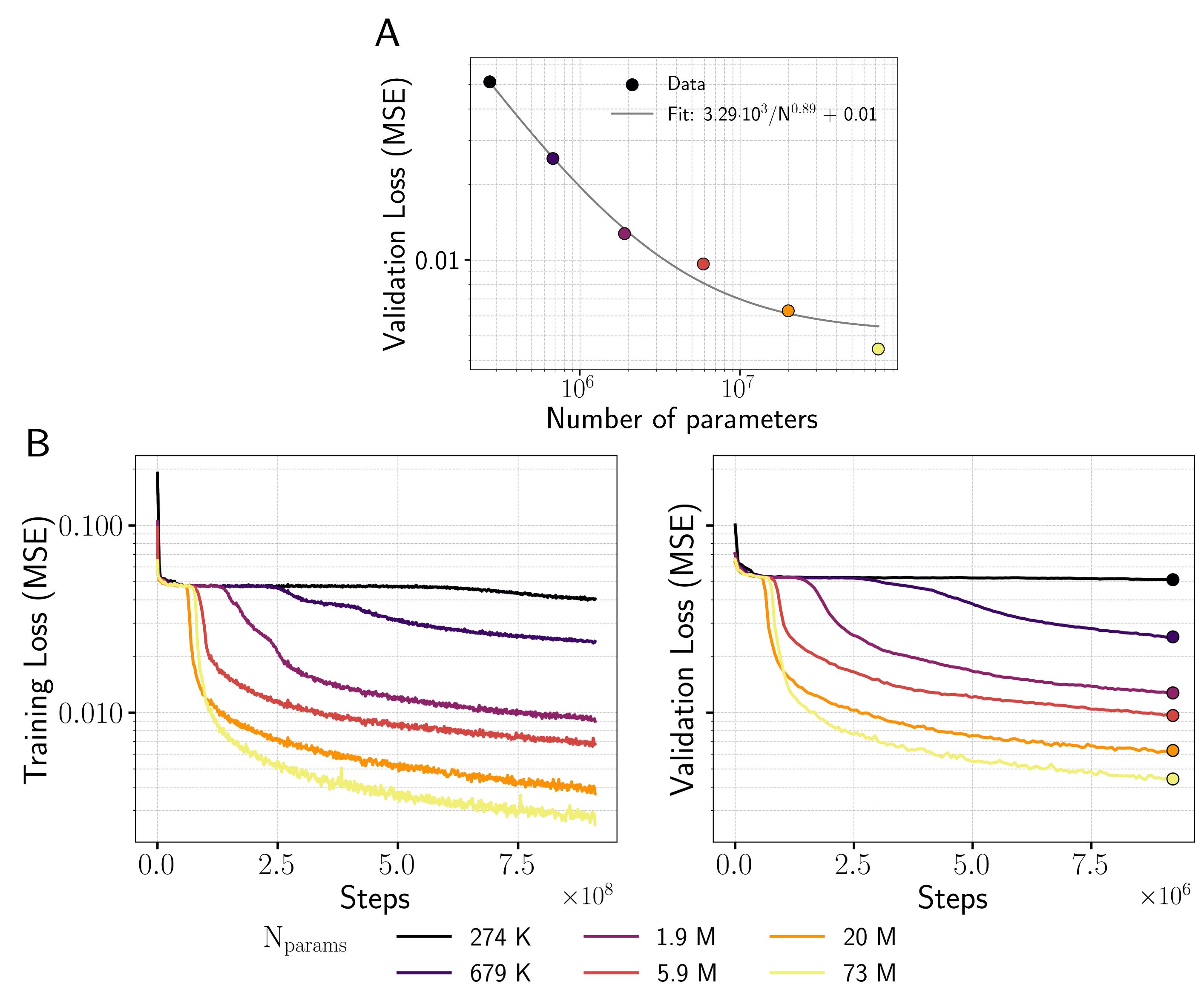}
    \caption{Scalability analysis of model performance with respect to parameter count. (A) Power-law fit of the validation loss as a function of the number of parameters, showing improved performance with increasing model size. (B) Training (left) and validation (right) loss curves for models of varying size, ranging from 274K to 73M parameters, plotted over training steps.}
    \label{fig:scalability}
\end{figure}

We conducted a series of experiments to evaluate how the number of parameters influences the performance of the model in predicting the failure pattern. 
The dataset used to train these models consists of an equal proportion of phase-field simulations and data generated by the rule-based algorithm. The use of both types of data was intended to create a complex learning scenario, suitable for evaluating model scalability.
In Figure~\ref{fig:scalability} we present the loss as a function of the model size, which ranges from 274k to 282M parameters. By increasing the model size, the final validation loss decreases.
To scale our model, we increased the number of channels in both the encoder and decoder, details about the scale-up strategy is presented in Table~\ref{tab:scaleup}. In Table~\ref{tab:hyperp}, the main architecture details and hyperparameters are summarized. The loss function employed in this work is the mean squared error, the optimizer is Adam~\cite{loshchilov2017decoupled}, and we used a linear warm-up of the learning rate  with a warm-up phase of 64000 steps.

\begin{table}[h]

\vspace*{1em}

\begin{tabular}{@{}lcccc@{}}
\toprule
$\mathrm{N_{params}}$ & \multicolumn{1}{c}{\begin{tabular}[c]{@{}c@{}}Number of channels\\  encoder\end{tabular}} & \multicolumn{1}{c}{\begin{tabular}[c]{@{}c@{}}Number of channels\\  decoder\end{tabular}} & Number of self attention layers &  \\ \midrule
\textcolor[HTML]{000000}{\fontsize{13pt}{13pt}\textbullet} 274\,146&    32   &   32     &   3    &  \\
\textcolor[HTML]{87216b}{\fontsize{13pt}{13pt}\textbullet} 679\,362&     64  &   64     &   3    &  \\
\textcolor[HTML]{ba3655}{\fontsize{13pt}{13pt}\textbullet} 1\,883\,010&    128   &   128     &   3    &  \\
\textcolor[HTML]{e45a31}{\fontsize{13pt}{13pt}\textbullet} 5\,863\,170&   256    &    256    &    3   &  \\
\textcolor[HTML]{f98c0a}{\fontsize{13pt}{13pt}\textbullet} 20\,114\,946&    512   &    512    &   3    &  \\
\textcolor[HTML]{f98c0a}{\fontsize{13pt}{13pt}\textbullet} 73\,784\,322 &    1024   &   1024     &   5    &  \\
\textcolor[HTML]{fcffa4}{\fontsize{13pt}{13pt}\textbullet} 281\,786\,370&    2048   &     2048   &    10   &  \\
\bottomrule
\end{tabular}
\vspace*{1em}
\caption{Scale-up of the models: hyperparameters tuned.}
\label{tab:scaleup}
\end{table}

\begin{table}[b]

\vspace*{1em}
\centering
\begin{tabular}{@{}lc@{}}

\toprule
Hyperparameter & Value \\ \midrule
Space bands in positional encoder &    32   \\
Number of latents &     2048  \\
Number of cross attention heads (encoder) &     2  \\
Number of self attention heads (encoder) &     2  \\
Number of cross attention heads (decoder) &     1  \\
Number of layers & Tab.~\ref{tab:scaleup} \\  
Number of channels & Tab.~\ref{tab:scaleup} \\
Gradient accumulation&     64  \\
Gradient clipping&     0.5  \\
Weight decay&     0.1  \\
Learning rate (set-point) & 10$^{-4}$ \\

\bottomrule
\end{tabular}

\vspace*{1em}
\caption{Hyperparameters of architecture and training}
\label{tab:hyperp}
\end{table}

\section{Surrogate model effect on performance}\label{sec:surr_model}
In this section, we expand on the results shown in Figure~\ref{fig:pre-pre-training}. Two approaches are compared: (i) training from scratch on the phase-field simulations, and (ii) a curriculum learning strategy involving pre-training on data generated using a rule-based algorithm, followed by fine-tuning on the phase-field simulations.
\begin{figure}[b]
    \centering
    \includegraphics[width=0.9\textwidth]{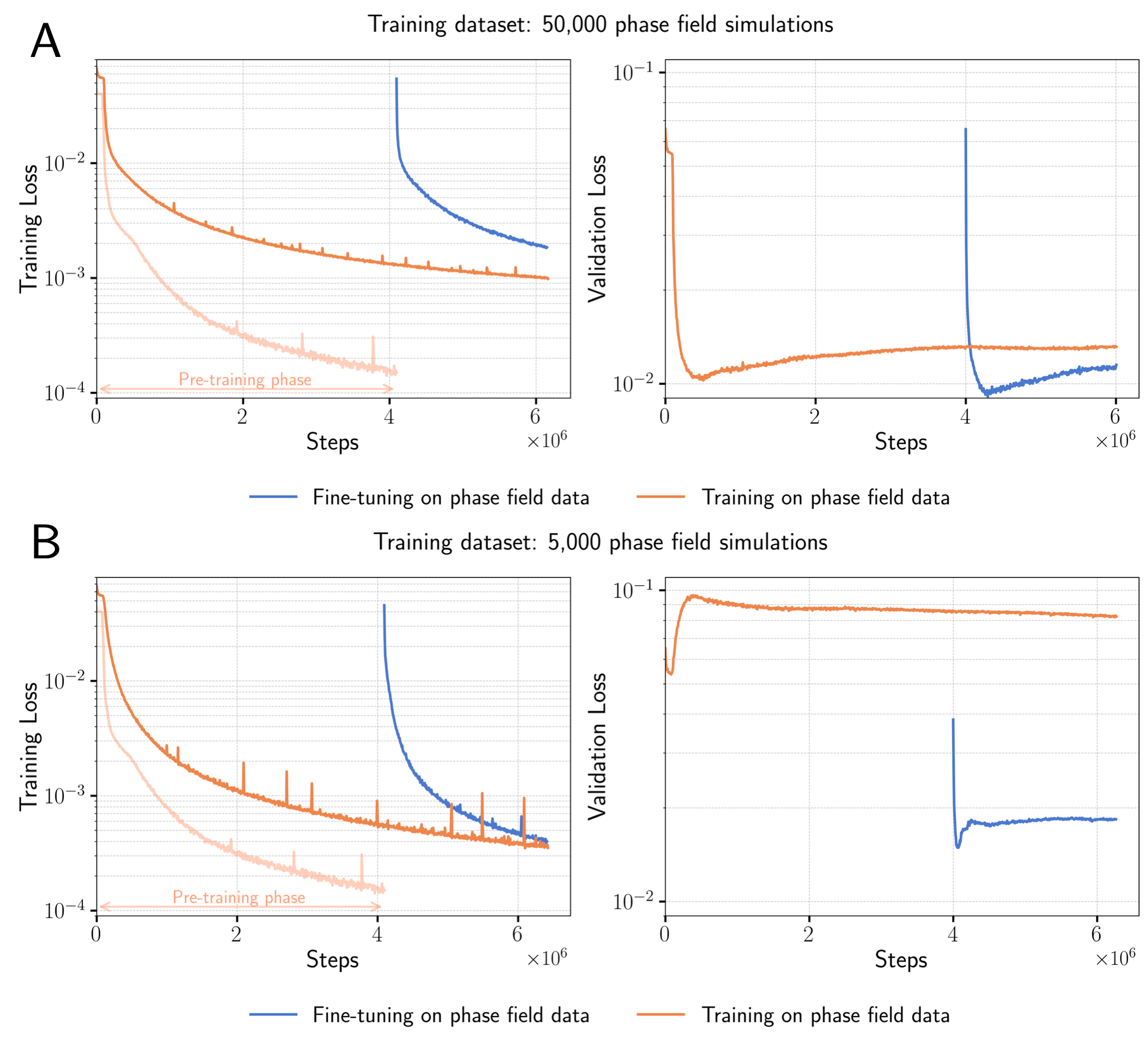}
    \caption{Training and validation loss for models trained with 50,000 (A) and 5,000 (B) phase-field simulations. Two approaches are compared: In blue, training from scratch directly on the phase-field simulations. In orange, a curriculum learning strategy is used: the pretraining phase (light orange) is performed on rule-based data, followed by fine-tuning (dark orange) on the phase-field simulations.}
    \label{fig:training-set-loss}
\end{figure}

Figure~\ref{fig:training-set-loss}A shows the training and validation curves when 5,000 phase-field simulations are used for training, while Figure~\ref{fig:training-set-loss}B presents the same plots for the case with 50,000 simulations. In both cases, the comparisons between the two approaches are made using the same number of training steps.

The analysis of the validation loss reveals that, when the training set is smaller, the model trained from scratch generalizes poorly—indicating overfitting to the training data. When the training set size increases to 50,000 samples, the performance of the two approaches becomes comparable. We can conclude that the curriculum learning strategy is significantly more effective in data-scarce regimes, which is a common scenario in scientific applications.

\section{Additional predictions}\label{sec:add_pred}
\begin{figure}[h]
    \centering
    \includegraphics[width=\textwidth]{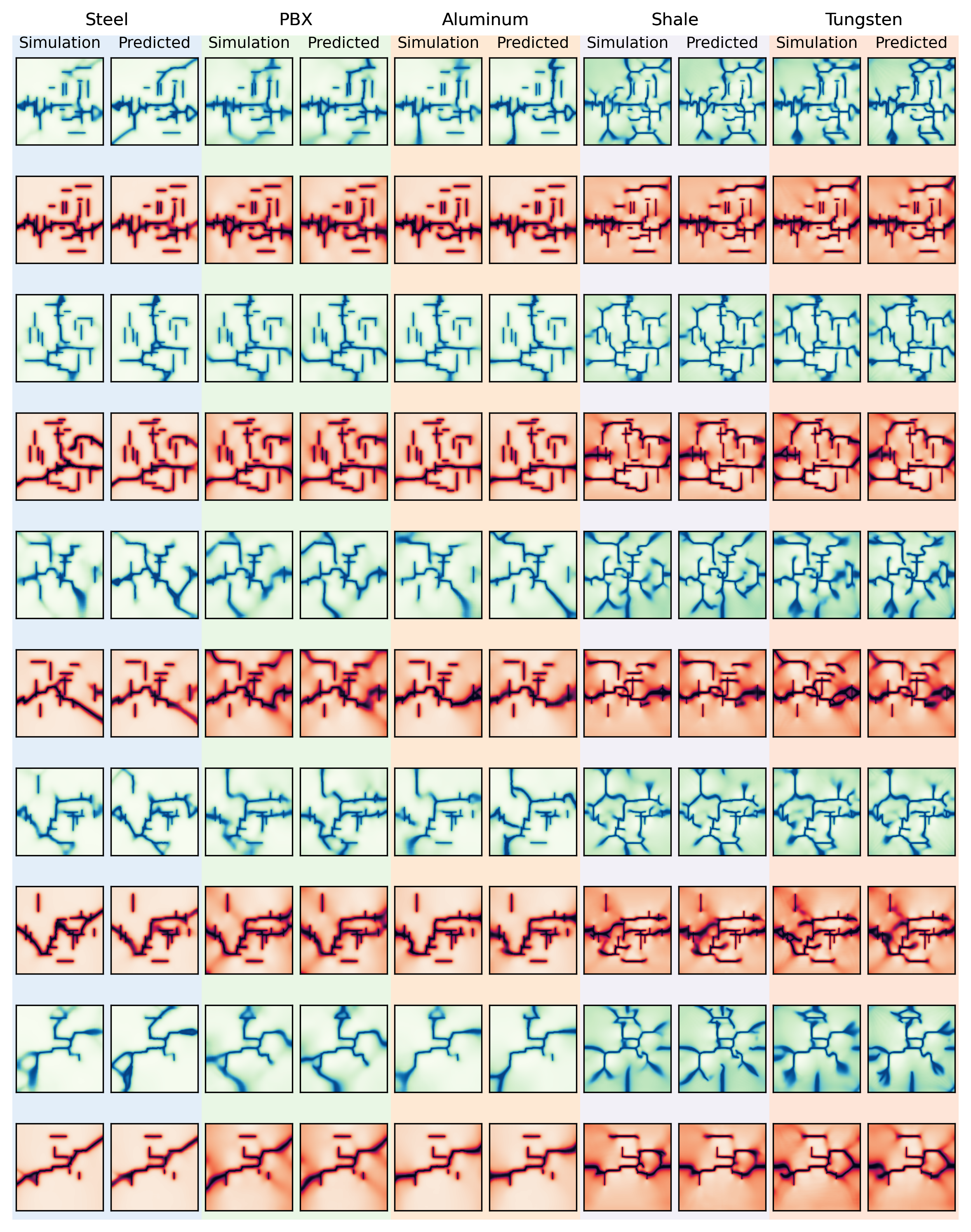}
    \caption{
    Additional predictions for varying orthogonal fracture initializations across materials and loading conditions. Predictions are shown for five different materials (steel, PBX, aluminum, shale, and tungsten) under two loading conditions: axial and biaxial tension. Each pair of rows corresponds to the same initial configuration subjected to the two loading conditions. Columns compare the simulation output (left) and model prediction (right). Colorbars are normalized from 0 to 1.}
    \label{fig:mosaic_orthogonal}
\end{figure}

\begin{figure}[h]
    \centering
    \includegraphics[width=0.9\textwidth]{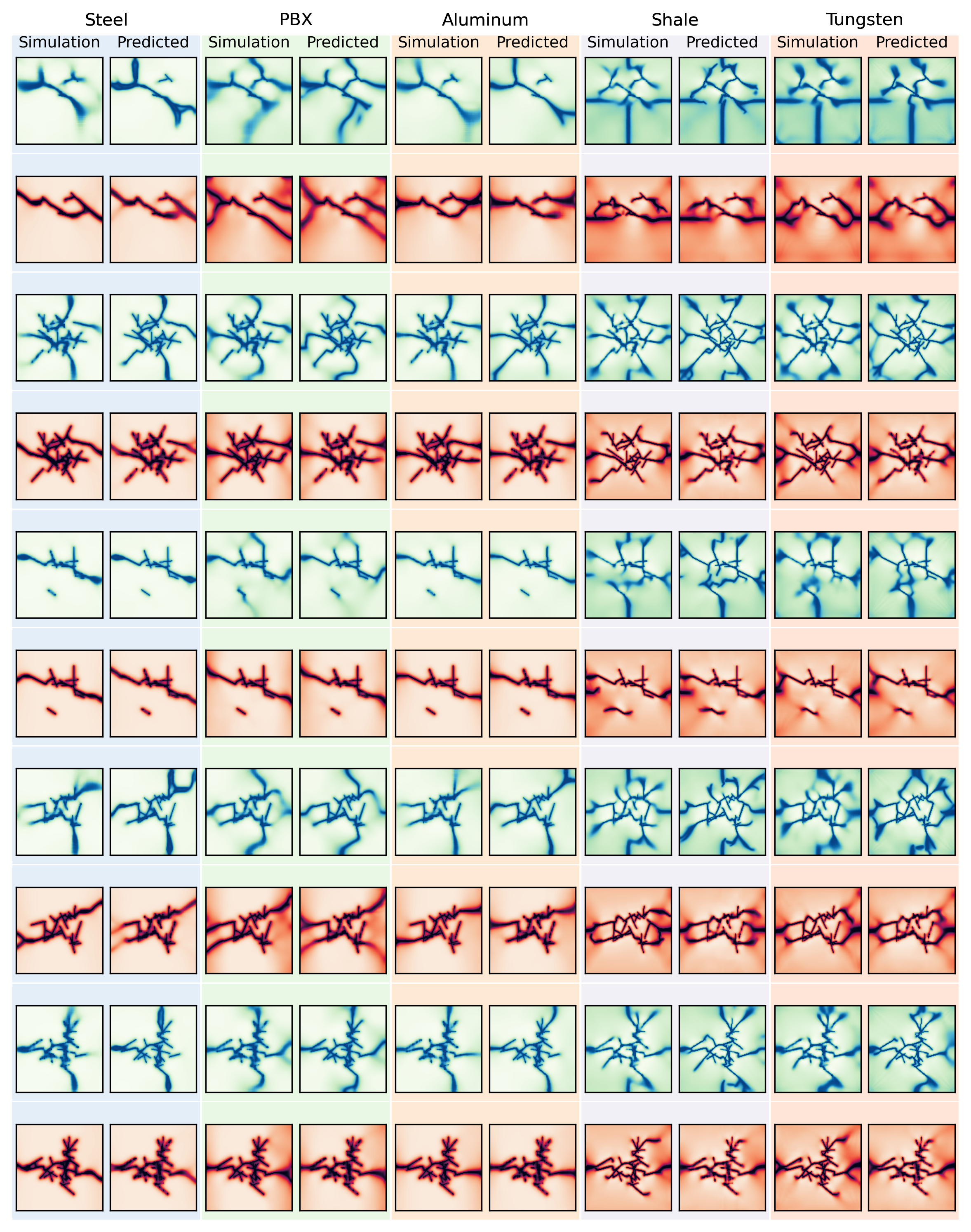}
    \caption{Additional predictions for varying random fracture initializations across materials and loading conditions. Predictions are shown for five different materials (steel, PBX, aluminum, shale, and tungsten) under two loading conditions: axial and biaxial tension. Each pair of rows corresponds to the same initial configuration subjected to the two loading conditions. Columns compare the simulation output (left) and model prediction (right). Colorbars are normalized from 0 to 1.}
    \label{fig:mosaic_random}
\end{figure}

In this section, we collect more examples of comparison between predicted and ground truth failure patterns. This model is trained on a phase-field dataset of 2,000,000 simulations, the training set contains orthogonal fracture initializations, we tested the generalization of our model to random fracture initialization. In Figure~\ref{fig:mosaic_orthogonal}, 50 predictions (from the test set) are qualitatively compared to the phase-field ground truth for orthogonal fracture initialization, in Figure~\ref{fig:mosaic_random}, 50 predictions are showed for random fracture initialization.  

\clearpage

\section{Test set errors}
The following tables summarize the model’s test set performance across materials and boundary conditions, evaluated on two generalization tasks: (1) prediction of failure time from initial fracture configurations with varying geometric characteristics, and (2) temporal prediction of intermediate fracture states. Table~\ref{tab:mae_results_generalization} reports the mean absolute error (MAE) across different types of initial conditions, including training-like samples, high or low fracture density, and randomized orientations. Table~\ref{tab:mae_results_time} shows the MAE for predicting fracture patterns at specific points in the failure progression. Results are broken down by material (steel, aluminum, PBX, shale, tungsten) and loading condition (axial and biaxial).

\section{Generalization to concrete}\label{sec:concrete}
\begin{figure}
    \centering
    \includegraphics[width=\textwidth]{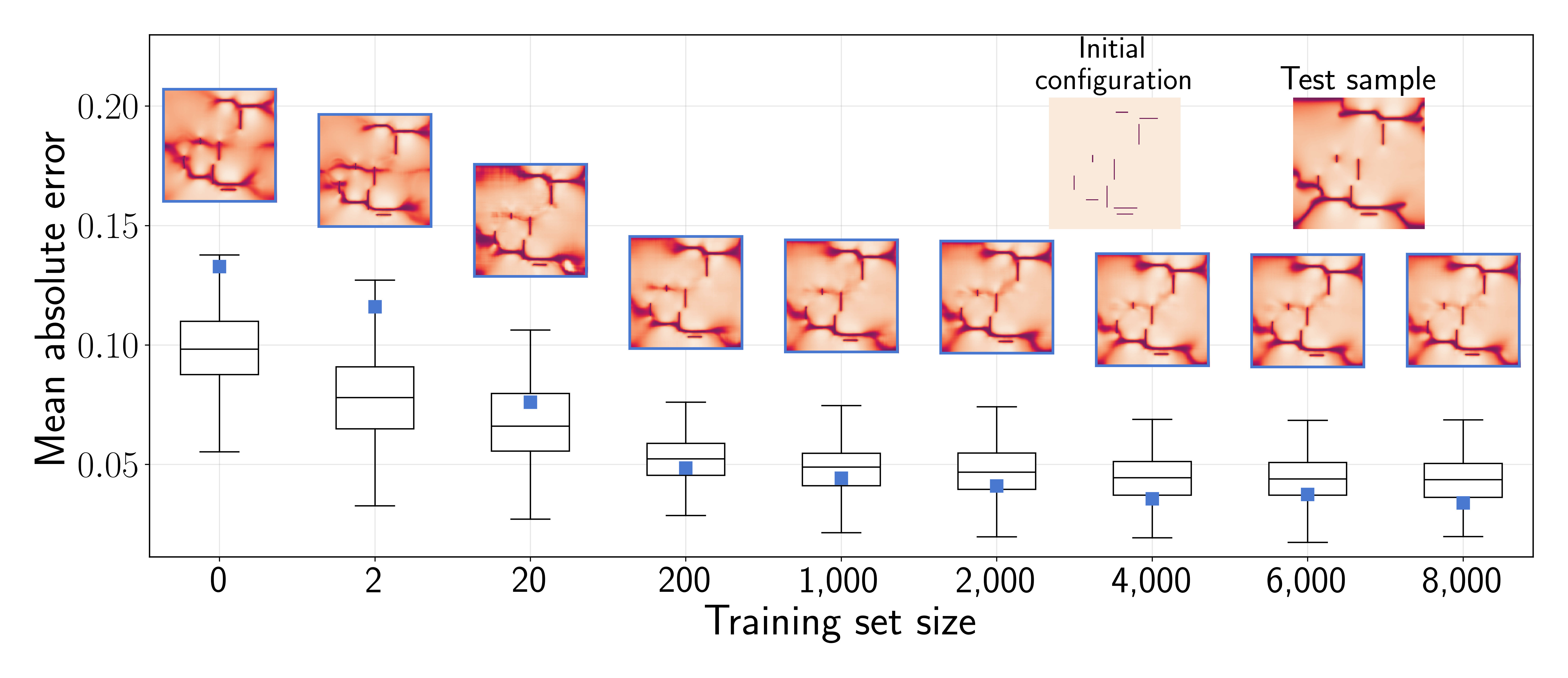}
    \caption{Failure pattern predictions for concrete (new material, not present in the base model training set)  as a function of the training set size.}
    \label{fig:fine-tuning-new-material-si}
\end{figure}

Figure~\ref{fig:fine-tuning-new-material-si} illustrates the fine-tuning performance of the foundation model on concrete, a material not included in the base pretraining dataset. The boxplots show the distribution of the mean absolute error as a function of the training set size, while the insets above each box depict the failure pattern predictions (top) for a selected test sample. The initial configuration and ground truth are shown in the top right corner for context. Without fine-tuning (0 samples), the model produces only coarse approximations of the failure pattern. As training set size increases, the predictions improve steadily, with a noticeable reduction in MAE and better capture of fracture propagation. From 20 samples, the model achieves satisfactory accuracy, with predictions closely matching the ground truth. This indicates that for a material with failure behavior different from the training data, effective generalization can be achieved with a relatively small number of examples. 

\begin{table}[htbp]
\centering
\caption{MAE on the test set for generalization to in- and out-of-distribution initial fracture configurations. Each column corresponds to a different type of initial configuration: training-like, high fracture density, single fracture, and random orientations. Each row reports results for a different material and loading condition.}
\label{tab:mae_results_generalization}

\begin{tabular}{l lcccc}
\toprule
\textbf{Material} & \makecell{\textbf{Loading}\\\textbf{Condition}} &
\makecell{\textbf{Training frac.}\\\textbf{density}} &
\makecell{\textbf{High frac.}\\\textbf{density}} &
\makecell{\textbf{Low frac.}\\\textbf{density}} &
\makecell{\textbf{Random frac.}\\\textbf{configuration}} \\
\midrule
\multirow{2}{*}{Steel}     
  & biaxial   & 0.049 & 0.064 & 0.590 & 0.046 \\
  & \cellcolor{lightgray}axial & \cellcolor{lightgray}0.030 & \cellcolor{lightgray}0.049 & \cellcolor{lightgray}1.332 & \cellcolor{lightgray}0.033 \\
\multirow{2}{*}{Aluminum}  
  & biaxial   & 0.036 & 0.039 & 0.209 & 0.040 \\
  & \cellcolor{lightgray}axial & \cellcolor{lightgray}0.022 & \cellcolor{lightgray}0.032 & \cellcolor{lightgray}0.232 & \cellcolor{lightgray}0.028 \\
\multirow{2}{*}{PBX}       
  & biaxial   & 0.034 & 0.055 & 0.439 & 0.043 \\
  & \cellcolor{lightgray}axial & \cellcolor{lightgray}0.026 & \cellcolor{lightgray}0.038 & \cellcolor{lightgray}0.495 & \cellcolor{lightgray}0.033 \\
\multirow{2}{*}{Shale}     
  & biaxial   & 0.031 & 0.157 & 0.519 & 0.051 \\
  & \cellcolor{lightgray}axial & \cellcolor{lightgray}0.023 & \cellcolor{lightgray}0.129 & \cellcolor{lightgray}0.490 & \cellcolor{lightgray}0.032 \\
\multirow{2}{*}{Tungsten}  
  & biaxial   & 0.031 & 0.168 & 0.551 & 0.056 \\
  & \cellcolor{lightgray}axial & \cellcolor{lightgray}0.026 & \cellcolor{lightgray}0.130 & \cellcolor{lightgray}0.562 & \cellcolor{lightgray}0.037 \\
\midrule
\multicolumn{2}{l}{\textbf{Mean}} &
\textbf{0.031} & \textbf{0.086} & \textbf{0.542} & \textbf{0.040} \\
\bottomrule
\end{tabular}
\end{table}

\begin{table}[htbp]
\centering
\caption{MAE on the test set for generalization to temporal predictions. Each row reports results for a different material and loading condition.}
\label{tab:mae_results_time}
\begin{tabular}{l l l}
\toprule
\textbf{Material} & \textbf{Loading Condition} & \textbf{MAE} \\
\midrule
\multirow{2}{*}{Steel}    
    & biaxial   & 0.0208 \\
    & \cellcolor{lightgray}axial & \cellcolor{lightgray}0.0175 \\
\multirow{2}{*}{Aluminum} 
    & biaxial   & 0.0196 \\
    & \cellcolor{lightgray}axial & \cellcolor{lightgray}0.0162 \\
\multirow{2}{*}{PBX}      
    & biaxial   & 0.0268 \\
    & \cellcolor{lightgray}axial & \cellcolor{lightgray}0.0289 \\
\multirow{2}{*}{Shale}    
    & biaxial   & 0.0314 \\
    & \cellcolor{lightgray}axial & \cellcolor{lightgray}0.0297 \\
\multirow{2}{*}{Tungsten} 
    & biaxial   & 0.0308 \\
    & \cellcolor{lightgray}axial & \cellcolor{lightgray}0.0350 \\
\midrule
\multicolumn{2}{l}{\textbf{Mean}} & \textbf{0.0257} \\
\bottomrule
\end{tabular}
\end{table}

\section{Data}\label{sec:data}

The foundation model presented in this work has been trained on different sources of simulation data. The base model has been trained on computationally inexpensive rule-based data generated on-the-fly during the training. The next training phase has focused on more expensive simulation data. In this section, we give an high level overview of the data. 

\subsection{Surrogate simulations: Rule-based model}
\label{sec:fwb}

\begin{figure}
    \makebox[\textwidth][c]{
        \includegraphics[width=1.3\textwidth]{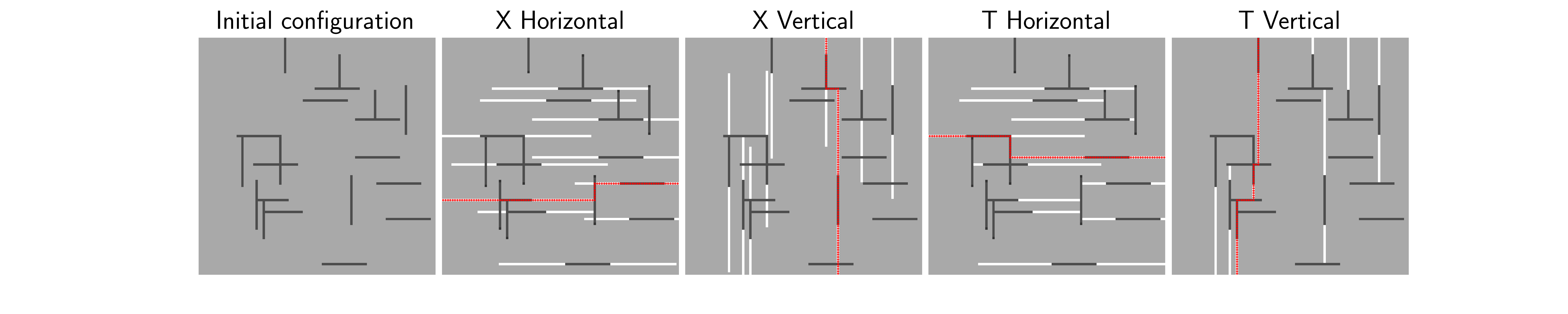}
    }
    \caption{Examples of outcomes of the rule-based algorithm. Initial configuration of the fractures, Failure pattern for the four different simulation modes: X horizontal, X vertical, T horizontal, T vertical. In red the path connecting the boundaries of the material (right-left for horizontal growth, top-bottom for vertical growth) causing the material failure.}
    \label{fig:wannabe}
\end{figure}

A rule-based algorithm simulates fracture propagation on a 2D material grid, starting from an initial configuration of fractures. Fractures grow step by step in either horizontal or vertical directions. In the T growth mode, they stop when they intersect with other fractures, while in the X mode, they pause temporarily before continuing. The grid updates over time, and material failure is detected by checking whether a continuous fracture path forms across the domain using graph-based analysis. Example failure patterns for different growth modes and directions are shown in Figure~\ref{fig:wannabe}.

The simulation is computationally efficient and allows for on-the-fly data generation during model pretraining. Generating diverse examples continuously helps improve generalization and reduces overfitting.

This algorithm simulates the propagation of fractures in a 2D material grid over time and checks for material failure. It begins with initializing a material matrix where cells represent non-fractured (0) or fractured (1) points. Fractures propagate from specified initial points (called ``tips''), which move in a predefined direction each timestep (horizontal or vertical). The fracture tips have a freezing mechanism, which temporarily halt their movement when they encounter other fractures, depending on the simulation mode. In the T growth mode, the fractures stop their expansion when they encounter other fractures, while in the X growth mode, the fractures freeze temporarily then continue to propagate in the same direction. As fractures propagate, the material grid is updated to reflect fractured cells over multiple time steps. The fractures are then analyzed using graph theory, where the grid is represented as a graph, and neighboring fractured cells are connected by edges. The algorithm checks if there is a path across the material, either vertically or horizontally, which would indicate material failure. If such a path exists, the simulation records the time of failure. Failure patterns for X or T, horizontal or vertical, are shown in Figure~\ref{fig:wannabe}. 

The generation of data via this rule-based algorithm is inexpensive, allowing for a robust pre-training of the model generating data on-the-fly without the need of storing and loading data.

\subsection{Phase-field simulations}
\label{sec:phasefield}

To generate high-resolution data for training and evaluating fracture prediction models, we employed a dynamic phase-field formulation that captures the nucleation, growth, and coalescence of fractures in two-dimensional solids. The phase-field method represents cracks as a smoothly varying scalar damage field $\psi \in [0,1]$, where $\psi = 0$ corresponds to intact material and $\psi = 1$ indicates fully broken regions. This scalar field is coupled to the mechanical response of the solid through a strain energy degradation mechanism, and it evolves over time in response to the elastic energy stored in the material.

The governing system of partial differential equations consists of two coupled equations. The displacement field $\bf{u}$ evolves according to the momentum conservation law in a damage-degraded medium:
\begin{equation}
    \rho \, \partial_t^2 \bf{u} = \nabla \cdot \boldsymbol{\sigma} + \bf{f},
\end{equation}
where $\rho$ is the mass density, $\bf{f}$ is the body force (taken as zero in our case), and the Cauchy stress tensor $\boldsymbol{\sigma}$ is defined as:
\begin{equation}
    \boldsymbol{\sigma}(\bf{u}, \psi) = \left[ (1 - \eta)(1 - \psi)^2 + \eta \right] \, \bf{C} : \varepsilon,
\end{equation}
with $\varepsilon = \frac{1}{2}(\nabla \bf{u} + \nabla \bf{u}^\top)$ the small-strain tensor, $\bf{C}$ the fourth-order elasticity tensor, and $\eta = 10^{-6}$ a numerical regularization constant that prevents full stiffness loss in fully damaged regions.

The evolution of the phase-field variable $\psi$ is driven by a balance between the elastic energy release rate and the regularized surface energy of the fracture:
\begin{equation}
    G_c \left( \frac{1}{w_0} \psi - w_0 \nabla^2 \psi \right) = 2(1 - \psi) \, H^+(\bf{u}),
\end{equation}
where $G_c$ is the critical energy release rate, $w_0$ is a length-scale parameter controlling the width of the diffused fracture zone, and $H^+(\bf{u})$ is a history variable storing the maximum tensile strain energy over time:
\begin{equation}
    H^+(\bf{u}, t) = \max_{\tau \in [0, t]} \psi^+(\varepsilon(\bf{u}(\tau))),
\end{equation}
with $\psi^+$ the positive part of the elastic energy density (associated with tension). This ensures that fracture growth is irreversible and driven by tensile loading only.

The spatial domain is discretized using $127 \times 127$ regular first-order spectral elements, resulting in $128 \times 128$ degrees of freedom for the scalar phase field and $2 \times 128 \times 128$ for the vector displacement field. Time integration is carried out using the implicit Newmark-beta method with parameters $\beta = 0.25$ and $\gamma = 0.5$, and the coupled system is solved using a staggered strategy.

Initial configurations are generated by randomly placing a number of horizontal and vertical fractures in the domain, with independent uniform sampling of their number (3–15 per direction), length (1–5~cm), and positions. Fracture apertures are sampled logarithmically between 0.5 and 5~mm. The domain is square, with side length 0.25~m. Fractures that intersect boundaries are clipped to fit within the domain.

Each configuration is simulated under two loading conditions:
\begin{itemize}
    \item \textbf{Uniaxial loading}: constant velocity $u_z^\pm = \pm t$ at the top and bottom boundaries (Dirichlet),
    \item \textbf{Biaxial loading}: simultaneous velocities in both directions, $u_x^\pm = \pm t$, $u_z^\pm = \pm t$.
\end{itemize}

The dataset spans five material systems: PBX, shale, tungsten, aluminum, and steel. PBX and shale are modeled as brittle elastic materials. Shale is implemented as a transversely isotropic medium, using its full elastic stiffness tensor in Voigt notation. Aluminum, steel, and tungsten are modeled as ductile materials using a rate-independent elastoplastic formulation. Yielding is modeled using the von Mises criterion:
\begin{equation}
    f = \sqrt{\frac{3}{2} \boldsymbol{s} : \boldsymbol{s}} - \sigma_y,
\end{equation}
where $\boldsymbol{s}$ is the deviatoric stress and $\sigma_y$ is the yield strength. Plastic deformation is governed by an associative flow rule and includes isotropic hardening. At each time step, the plastic strain is updated and the consistent tangent stiffness is assembled accordingly.

The physical parameters used for each material are listed in Table~\ref{tab:pf_materials}.

\begin{table}[h]
    \centering
    \caption{Material parameters used in the phase-field simulations.}
    \label{tab:pf_materials}
    \small
    \renewcommand{\arraystretch}{1.2}
    \begin{tabular}{lccccc}
        \toprule
        \textbf{Property} & \textbf{PBX} & \textbf{Shale} & \textbf{Tungsten} & \textbf{Aluminum} & \textbf{Steel} \\
        \midrule
        Density ($\times 10^3$~kg/m$^3$) & 1.82 & 2.075 & 19.25 & 2.7 & 7.85 \\
        Young’s modulus (GPa) & 10 & — & 400  & 64.9 & 200 \\
        Poisson’s ratio & 0.36 & — & 0.28 & 0.25 & 0.30 \\
        $C_{11}$ (GPa) & — & 31.3 & — & — & — \\
        $C_{13}$ (GPa) & — & 3.40 & — & — & — \\
        $C_{33}$ (GPa) & — & 22.5 & — & — & — \\
        $C_{55}$ (GPa) & — & 6.49 & — & — & — \\
        Yield strength (GPa) & — & — & 0.75 & 0.25 & 0.6 \\
        Hardening modulus (GPa) & — & — & 5.0 & 0.25 & 2.5 \\
        $G_c$ (J/m$^2$) & 641 & 50 & 500 & $1 \times 10^4$ & $2.5 \times 10^5$ \\
        \bottomrule
    \end{tabular}

    \begin{tabular}{lcc}
        \toprule
        \textbf{Property} & \textbf{Titanium} & \textbf{Concrete}\\
        \midrule
        Density ($\times 10^3$~kg/m$^3$) & 4.43 & 2.4 \\
        Young's modulus (GPa) & 115 & 30  \\
        Poisson's ratio &0.33 &0.15 \\
        Yield strength (GPa) & 1.0 & - \\
        Hardening modulus (GPa) & 2.0 & - \\
        $G_c$ (J/m$^2$) &$2.4\times 10^4$ & 150\\
        \bottomrule
    \end{tabular}

\end{table}

Each simulation runs until a failure criterion is met, defined as $\max(\psi) > 0.99$. For each case, we extract: (i) the initial configuration (fracture mask and metadata), (ii) the final fracture pattern (binary threshold of $\psi$), and (iii) the failure time. This results in a dataset of over 400,000 simulations spanning different materials, initial conditions, and loading regimes, and provides the foundation for training and evaluation throughout this work.

\subsection{Combined finite-discrete element simulations}
\label{sec:hoss-pbx}

HOSS employs the combined finite-discrete element method \cite{munjiza_combined_2004}, integrating finite element analysis of continuous media with discrete element techniques for transient dynamics and contact interactions.  
The solid domain is discretized into finite elements supporting finite rotations, large displacements through multiplicative decomposition \cite{munjiza_large_2015,lei2016generalized}, and selective stress integration to prevent artificial stiffness \cite{lei2016non}.  
A unified hypo/hyper-elastic formulation allows user-defined isotropic and anisotropic material models \cite{lei2016generalized}.  
During failure and fragmentation, single finite element meshes transition into multiple interacting domains using the same finite element framework for contact handling, with appropriate discretization methods managing contact detection and interactions \cite{munjiza_compmechdis_2011,Knight_etal_2020,lei2014framework}.

To generate our fracture simulation data we incorporate initial seeded fractures developed by methods described in Section~\ref{sec:phasefield}. These randomly distributed fractures populate a 2D square domain measuring 0.25 m by 0.25 m, confined by 5 mm plates.  
The model employs free boundary conditions along the sides while applying linear ramping nodal velocity boundary conditions ($\pm$1 m/s) in extension to the plates over 0 to $1\cdot10^{-4}$ s. This gradual boundary condition approach prevents unwanted material breakage at the sample boundaries due to sudden loading enforcements. Simulations are constrained by a fixed time increment of $1\cdot10^{-8}$ s and run for approximately 8 hours using 8 MPI domains and CPU cores to capture complete material failure.

We investigate PBX, consistent with the material used in the phase-field simulations. PBX represents brittle elastic fracture behavior common to geo-materials including rocks and concrete, exhibiting quasi-brittle characteristics with localized damage preceding complete fracture.  

A systematic choice of the time-to-failure requires careful consideration of key indicators to accurately determine material failure.  
Time-to-failure results are obtained by post-processing unique HOSS-tracked variables that do not directly influence final fracture patterns.  
HOSS tracks the entire solid kinetic energy of the system and also the stress tensor for each element.  
Time-to-failure is determined using an automated picking routine: (1) evaluate total stress and kinetic energy at each time step; (2) identify all stress and energy peaks during the simulation; (3) select the key kinetic energy peak with the steepest drop, marking failure onset; (4) locate the first stress peak after this energy drop, indicating the transition to failure; and (5) compute the stress gradient to identify when stress ceases decreasing, signaling final failure.  
This approach provides a systematic estimate for the time-to-failure.  
Visual inspection confirmed that this method aligned with near- or full-sample fracture propagation.  
Determination of material failure can vary, but our method provides a conservative estimate as actual material failure may occur earlier if factors like sudden stress drops are considered.
The physical parameters used for PBX are listed in Table~\ref{tab:materials_HOSS_PBX}.
\begin{table}[h]
    \centering
    \caption{Material properties for HOSS simulations}
    \label{tab:materials_HOSS_PBX}
    \begin{tabular}{lc}
        \toprule
        \textbf{Property} & \textbf{PBX} \\
        \midrule
        Density ($\times 10^3$~kg/m$^3$) & 1.82 \\
        Young's Modulus (GPa) & 10  \\
        Poisson's ratio & 0.36 \\
        Coulomb Friction & 0.6 \\
        Tensile Strength (GPa) & 0.0005 \\
        Shear Strength (GPa) & 0.002 \\
        $G_{I}$ (J/m$^2$) & 125 \\
        $G_{II}$ (J/m$^2$) & 500 \\
        \bottomrule
    \end{tabular}
\end{table}

\clearpage
\section*{Data availability}
The data is available at \url{https://huggingface.co/datasets/smartFRACs/material_fracturing}.
\section*{Acknowledgments} 
This research work was supported by the Laboratory Directed Research and Development program of Los Alamos National Laboratory under project number 20250637DI.
Los Alamos National Laboratory is operated by Triad National Security, LLC, for the National Nuclear Security Administration of U.S. Department of Energy (Contract No. 89233218CNA000001). DO gratefully acknowledges partial support from the Department of Energy, Office of Science, Office of Basic Energy Sciences, Geoscience Research program under Award Number LANLECA1.

\section*{Author contributions}

A.M.: Formal analysis, Investigation, Writing – original draft, Visualization, Methodology; A.P.: Formal analysis, Investigation, Writing – original draft, Visualization, Methodology; R.G.H: Formal analysis, Investigation, Data curation, Writing - original draft, Visualization; K.G.: Data curation, Writing - original draft; X.W.: Data curation; E.R.: HOSS technical advice and guidance, Writing - review \& editing, Z.L.: HOSS technical advice and guidance, V.A.: Writing – original draft; J.C.: Writing - review \& editing; Q.K.: Writing - review \& editing; J.D.H.: Writing - review \& editing; A.H.: Writing - review \& editing; N.D.: Writing - review \& editing, Resources, Funding acquisition; E.L.: Writing - review \& editing, Resources, Funding acquisition, Conceptualization;  H.V.: Conceptualization, Writing - review \& editing, Funding acquisition; D.O.: Conceptualization, Methodology, Software, Writing - review \& editing, Funding acquisition; J.E.S.: Conceptualization, Methodology, Software, Writing – original draft.

\bibliography{sn-bibliography}

\end{document}